\documentclass[10pt,twocolumn,letterpaper]{article}

\usepackage{cvpr}              

\usepackage[dvipsnames]{xcolor}

\usepackage{textcomp}  
\usepackage{csquotes}  
\usepackage{glossaries}
\newacronym{nerf}{NeRF}{Neural Radiance Field}
\newacronym{mlp}{MLP}{multilayer perceptron}
\newacronym{ad}{AD}{anomaly detection}
\newacronym{mad}{MAD}{multi-pose anomaly detection}
\newacronym{3dgs}{3DGS}{3D Gaussian Splatting}
\newacronym{gan}{GAN}{generative adversarial network}
\newacronym{mvs}{MVS}{multi-view stereo}
\newacronym{sfm}{SfM}{structure-from-motion}
\newacronym{sh}{SH}{spherical harmonic}
\newacronym{ssim}{SSIM}{structural similarity index measure}
\newacronym{loftr}{LoFTR}{Local Feature Matching with Transformers}

\DeclareMathOperator*{\argmin}{argmin} 
\usepackage{float}

\usepackage{placeins}
\usepackage{stfloats}
\usepackage{graphicx}
\usepackage{amsmath}
\usepackage{amssymb}
\usepackage{booktabs}
\usepackage{bm}
\usepackage{tabularx}
\newcolumntype{C}{>{\centering\arraybackslash}X}
\NewExpandableDocumentCommand\mcc{O{1}m}{\multicolumn{#1}{c}{#2}}

\usepackage[accsupp]{axessibility}
\usepackage{multirow}
\usepackage{svg}

\definecolor{cvprblue}{rgb}{0.21,0.49,0.74}
\usepackage[pagebackref,breaklinks,colorlinks,citecolor=cvprblue]{hyperref}

\def\paperID{***} 
\def\confName{CVPR}
\def\confYear{2024}

\title{SplatPose \& Detect: Pose-Agnostic 3D Anomaly Detection}

\author{Mathis Kruse, Marco Rudolph, Dominik Woiwode, Bodo Rosenhahn\\
Institute for Information Processing, 
Leibniz University Hannover\\
{\tt\small \{kruse, rudolph, woiwode, rosenhahn\}@tnt.uni-hannover.de}
}

\begin{document}
\maketitle

\def\method{\emph{SplatPose}}
\def\methodreg{SplatPose}

\begin{abstract}
Detecting anomalies in images has become a well-explored problem in both academia and industry.
State-of-the-art algorithms are able to detect defects in increasingly difficult settings and data modalities.
However, most current methods are not suited to address 3D objects captured from differing poses.
While solutions using \glspl{nerf} have been proposed, they suffer from excessive computation requirements, which hinder real-world usability.
For this reason, we propose the novel 3D Gaussian splatting-based framework \method{} which, given multi-view images of a 3D object, accurately estimates the pose of unseen views in a differentiable manner, and detects anomalies in them.
We achieve state-of-the-art results in both training and inference speed, and detection performance, even when using less training data than competing methods.
We thoroughly evaluate our framework using the recently proposed Pose-agnostic Anomaly Detection benchmark and its \gls{mad} data set.
\end{abstract}

\section{Introduction}
\label{sec:intro}
In the industrial manufacturing of products, small errors and slight deviations from the norm inevitably lead to some defective products.
To ensure both quality and effectiveness of the production process, the detection of defects or irregularities becomes necessary.
Traditionally, this is done by a human supervisor trained in recognizing the special kinds of defects that may appear.
In most cases, this is far more costly and prone to human error than fully automated solutions.
Data of different kinds of defects are however scarcely available.
This unavailability of anomalous samples leads standard classification approaches to underperform in these settings.
More recently anomaly detection algorithms have shifted to train on normal data only~\cite{MVTecAD}, which is usually more widely available.

\begin{figure}
    \centering
    \includegraphics[width=\linewidth]{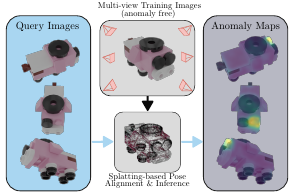}
    \caption{Example of \methodreg{}. A cloud representation is built from multi-view training images. During inference, query images with unknown poses are aligned and an anomaly map localizes their defects within the 3D object, irrespective of pose.}
    \label{fig:teaser}
\end{figure}

While state-of-the-art methods already perform very well on image data~\cite{AD:patchcore, MVTecAD}, more challenging benchmarks have emerged~\cite{Mvtecloco, MVTEC3D}, with some including pose information~\cite{madsim}.
For this exact task, the \gls{mad} data set has been proposed~\cite{madsim}, consisting of different views of 20 custom-built LEGO\textregistered{} figures and including anomalies such as missing bricks, discolorations, and artifacts caused by excess material.
As standard anomaly detection approaches do not differentiate between poses, they underperform heavily in this setting~\cite{madsim}.
In practice, objects in production may be randomly placed or rotated, while cameras are not adjustable as freely.
This makes the \gls{mad} data set a good benchmark for the anomaly detection community to verify the robustness of their methods.

OmniposeAD has been proposed in conjunction with the \gls{mad} data set.
It captures the anomaly-free volume density within a \gls{nerf} and aligns test images with unknown poses using the iNeRF framework~\cite{madsim, iNERF}.
Despite working in principle, the high computational costs make it impractical for real-world usage.
Also, \glspl{nerf} are known to struggle in sparse-view settings~\cite{regnerf}, limiting their applicability.

Recent progress on novel-view synthesis using \gls{3dgs} showed that the volume density of complex three-dimensional objects can be encoded into a cloud of three-dimensional Gaussians using multi-view images and their respective camera poses~\cite{3DGS}.
Advances in this field enabled practitioners to generate realistic unseen views of complex objects while still achieving high frame rates suitable for real-time execution~\cite{3DGS}.

In this work, we propose~\method{} to tackle 3D pose-agnostic anomaly detection using \gls{3dgs}.
The explicit 3D point cloud representation combined with the efficient rasterization results in up to $\textbf{55}$ times faster training and $\textbf{13}$ times faster inference times than other top competitors.
We apply transformations to the 3D point cloud to perform pose estimation in a differentiable manner, letting us render defect-free images of arbitrary pose.
Irrespective of a query object's pose, anomalies can be detected by matching features between the aligned rendering and query image, as shown in~\cref{fig:teaser}.
\methodreg{} improves the anomaly detection performance on \gls{mad}, reaching a new state-of-the-art in both detection and segmentation.
Even when using only $60\%$ of training data, we are able to outperform all other methods, while they utilize the entire training set.
Our contributions may be summarized as follows:
\begin{itemize}
    \item We propose \method{} for pose-agnostic anomaly detection, using 3D Gaussian Splatting to perform pose estimation in a differentiable manner.
    \item We achieve new state-of-the-art results on pose-agnostic anomaly detection.
    \item The proposed method is significantly more resource-efficient, both in terms of speed and required data, than other top competitors.
    \item Code is available on GitHub\footnote{\url{https://github.com/m-kruse98/SplatPose}}.
\end{itemize}

\section{Related Work}\label{sec:related}
As this paper deals with performing anomaly detection on multi-view images using novel-view synthesis techniques, a rough overview is given for both fields.

\subsection{Industrial Anomaly Detection}\label{sec:ad}

In semi-supervised anomaly detection, also known as one-class classification or novelty detection, the task is to differentiate between anomalous and normal samples, while only learning from normal samples.

One line of work uses generative models like \glspl{gan} or Autoencoders, which, when trained only on normal data, fail to reconstruct anomalous regions, thus enabling anomaly detection~\cite{MVTecAD, AD:anogan, AD:OCGAN}.
Another line of work tries to model the problem using density estimation, by assigning high likelihoods to the normal samples they are trained on, while anomalies receive much lower likelihoods. 
Leveraging complex pre-trained feature embeddings, and assuming them to be distributed as multivariate Gaussians, the Mahalanobis distance has been used for this density estimation~\cite{AD:maha_gde, AD:padim}.
In order to forego this distributional assumption, normalizing flows have been used, as they can map an arbitrarily complex feature distribution into a tractable Gaussian distribution~\cite{AD:differnet, AD:csflow, AD:cflow-ad, quantum_nf}.
A proxy for the density can also be constructed with the nearest-neighbor distance, which is measured between a test sample and all normal features gathered from the training set and saved within a memory bank~\cite{AD:patchcore, AD:spade, AD:HybridFusion, AD:CFA}.
Many of these methods need to choose meaningful features from pre-trained networks, which in itself is not trivial~\cite{HecklerFeatures}. Student-Teacher architectures have also seen success~\cite{AD:AST, AD:STFPM, AD:RD4AD, AD:UninfST}, as the student only learns to mimic the teacher on normal data and the regression error is also used as a proxy for density. Lastly, anomaly detection can also be learned by constructing synthetic anomalies and casting the problem into a supervised classification problem~\cite{AD:cutpaste, AD:draem, AD:simplenet}. However, synthetic anomalies introduce a bias that may cause detection to fail on unseen anomalies. 

With current methods excelling at detecting anomalies in high-quality images, anomaly detection data sets have shifted their attention to more difficult problems~\cite{Mvtecloco, MVTEC3D}.
The \gls{mad} data set serves as another example of this, as images no longer constrain themselves to fixed viewpoints, making reasoning about the object's pose crucial~\cite{madsim}.

\subsection{Novel-View Synthesis}\label{sec:2_view_synth}

In computer vision, 3D reconstruction is the task of estimating the three-dimensional representation of an object or a scene given several 2D images \cite{multi-view-geom}.
More specifically, one may want to recover the camera parameters, their poses and a sparse (or dense) 3D scene geometry, i.e. in the form of a 3D point cloud or mesh structures.
Traditionally, \emph{\gls{sfm}} algorithms have been extensively studied in the literature \cite{SFM:Snavely06, SFM:colmap} for extracting precise camera poses and coarse 3D representations of increasingly complex scenes. 
These mostly leverage feature-matching techniques and geometric constraints between neighboring views.
With the results from \gls{sfm} as input, more dense and precise 3D models of the object are generated using \emph{\gls{mvs}} pipelines~\cite{MVS:colmap}.
Novel views can then be extracted from these representations.

More recently, research on learning-based novel-view synthesis has increased in exposure, due to the improvements in neural rendering techniques such as \glspl{nerf}~\cite{NERF}, which encode the volumetric density within the weights of a \gls{mlp}.
Other approaches, namely \gls{3dgs}~\cite{3DGS}, encode the scene as a 3D point cloud of 3D Gaussian distributions, and have achieved similar photo-realistic view synthesis of increasingly bigger and more complex scenes while yielding a more explicit scene representation.
These methods raised the state-of-the-art in view synthesis and spawned lots of follow-up research~\cite{MipNeRF360, plenoctrees, regnerf, nerf_humannerf, nerf_dnerf, iNERF}.
Since they are crucial to our method, we will explain them in more detail in the following sections.

\subsubsection{NeRF and iNeRF}

Given a set of input views with corresponding camera poses, \glspl{nerf}~\cite{NERF} optimize a \gls{mlp} with parameters $\Theta$, to represent a continuous volumetric scene function.
This function maps from 5D coordinates, containing the 3D spatial location $\bm{x}$ and the 2D viewing direction $\bm{d}$, to the emitted color $\bm{c}$ and volume density $\sigma$, i.e. $f_{\Theta}: \left(\bm{x}, \bm{d}\right) \mapsto \left(\bm{c}, \sigma\right)$.
In order to render a pixel, \glspl{nerf} march a ray $\bm{r}\left(t\right) = \bm{o} + t\bm{d}$ from a starting point $\bm{o}$ in the direction $\bm{d}$ with $t \in \left[t_n, t_f\right]$. The color $\bm{C}(\bm{r})$ emitted by this ray is calculated as
\begin{equation}\label{eq:rendering1}
    \bm{C}(\bm{r}) = \int_{t_n}^{t_f} \mathcal{T}(t)\cdot\sigma(\bm{r}(t))\cdot\bm{c}(\bm{r}(t),\bm{d})\,dt,
\end{equation}
where
\begin{equation}\label{eq:rendering2}
    \mathcal{T}(t) = \text{exp}\left(-\int_{t_n}^{t}\sigma(\bm{r}(s))ds \right)
\end{equation}
accumulates the probability of a ray marching from $t_n$ to $t$ without hitting another particle, and $\sigma(\bm{x})$ denotes the probability of a ray ending at a particle at location $\bm{x}$.
To estimate this integral numerically, points are sampled along these rays, and the \gls{mlp} predicts color $\bm{c}$ and density $\sigma$, optimized via the photometric loss, i.e. the total squared error between rendering and ground truth image.

Some of the downsides of \glspl{nerf} include their very high computational costs with training times of several hours per scene, and their performance problems in sparse-view settings~\cite{NERF, regnerf}.
Further improvements have been proposed to address issues such as poor rendering quality~\cite{MipNeRF360, regnerf} or high inference times~\cite{plenoctrees}, among others.

By being differentiable, \glspl{nerf} can be used for pose estimation of novel views by \enquote{\textit{inverting}} the trained \gls{nerf} $f_{\Theta}$, spawning the \mbox{iNeRF} framework~\cite{iNERF}. The unknown camera pose $\hat{T}$ of an image $I$, is recovered by keeping $\Theta$ frozen and optimizing the problem
\begin{equation}
    \hat{T} = \argmin_{T\in SE(3)} \mathcal{L}\left(T|I,\Theta\right),
\end{equation}
where $\mathcal{L}$ again measures the photometric loss between $I$ and the image rendered using the camera pose transformation $T$, with $T$ sampled from the group of euclidean motions $SE(3)$.
Both \gls{nerf} and iNeRFs are sensitive to hyperparameters to ensure convergence at a reasonable speed.
The computational intensity of iNeRF remains an open problem.

\subsubsection{3D Gaussian Splatting}

By encoding the scene using a 3D point cloud of Gaussians, \gls{3dgs}~\cite{3DGS} is able to synthesize images of very high quality, while being efficient enough to render in real-time.
Each 3D Gaussian in this point cloud is characterized by its center position $\mu$ and a covariance matrix $\Sigma$, which is constructed using a scaling matrix $S$ and a rotation matrix $R$, utilizing the equation $\Sigma = RSS^TR^T$.
Further parameters include an opacity factor $\alpha$ and a color component $c$, which is modeled using \glspl{sh}, that act as a view-direction-dependent coloring of the Gaussian sphere.

Initially, given a set of multi-view images, their camera poses and an initial point cloud are estimated using \gls{sfm} pipelines~\cite{SFM:colmap}.
During optimization, novel views are repeatedly rendered using an efficient differentiable tile-based rasterizer and compared to their ground truth image. 
Since the rendering process is differentiable, the parameters of the Gaussians can be optimized using the mean squared error as well as the \gls{ssim}~\cite{SSIM}.

The rendering process is, just as with \glspl{nerf}, done along rays analogously to~\cref{eq:rendering1,eq:rendering2}.
In the case for \gls{3dgs}, the densities $\sigma$ and color values $\mathbf{c}$ are not sampled by querying an \gls{mlp}, but rather saved explicitly within the Gaussian representations.
Rasterization, efficient sorting and $\alpha$-blending Gaussians at the queried pixels enables fast and high-quality image synthesis~\cite{3DGS}.
During optimization, the cloud of Gaussians is controlled and optimized in an adaptive rule-based manner.
Some strategies include pruning away any points with negligibly small opacity values, proposing new points in sparsely populated areas, and splitting high-variance Gaussians into two of lower variance.

\subsubsection{OmniposeAD}\label{sec:rel_omnipose}

Since our work is inspired by OmniposeAD~\cite{madsim} (in this paper shortened to OmniAD), we describe its pipeline in the following.
To become pose-agnostic, OmniAD trains a \gls{nerf}, just as described by Mildenhall \etal~\cite{NERF}, and subsequently uses iNeRF to perform pose estimation.

The iNeRF framework relies on an initial pose, to start optimization. 
In line with the author's reference implementation, the \enquote{\emph{\gls{loftr}}} framework~\cite{LOFTR} is used to collect feature matches between the test sample and every training sample. The training sample with the most matches is chosen as the initial coarse pose. 
Afterward, pose estimation for OmniAD follows the iNeRF~\cite{iNERF} framework. Repeatedly rendering images with a \gls{nerf} becomes a major bottleneck for fast inference and computational requirements. 

Lastly, with sufficiently good pose estimation, OmniAD generates an image using the trained \gls{nerf} at the estimated pose.
As the \gls{nerf} is trained only on defect-free data, it is void of any anomalies, apart from any \gls{nerf}-related rendering mistakes.
The synthesized image is then compared to the original query sample, by gathering a pyramid of features from a pre-trained neural network for both images and calculating their mean squared error at every pixel.
Aggregating these pixel-wise score maps is done for image-wise anomaly detection. 

\section{Method}\label{sec:method}

\begin{figure}
    \centering
    \includegraphics[width=\linewidth]{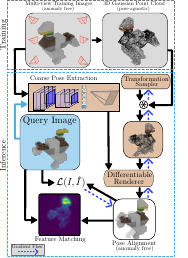}
    \caption{Overview of our pipeline. Multi-view training images are represented in a 3D point cloud of Gaussians. The unknown camera pose of a query image is first coarsely estimated and then iteratively refined by applying a pose transformation on the 3D point cloud before the differentiable renderer. The final anomaly-free rendering is then compared to the original test image to perform pixel-wise comparison for anomaly detection.}
    \label{fig:framework_overview}
\end{figure}

In the pose-agnostic anomaly detection setting, defect-free multi-view images and their respective poses are provided during training.
At inference time, the task is to infer whether a test sample contains any anomalies without having access to its camera pose.
We aim to solve this task using \method{}, which is visualized in~\cref{fig:framework_overview}.
We take inspiration from the OmniAD pipeline as described in~\cref{sec:rel_omnipose}, which solves the problem by coarsely estimating a test sample's pose from all training examples and then refining it with a previously optimized \gls{nerf} model.

Instead of using \glspl{nerf}, we resort to modeling the defect-free objects with \gls{3dgs} and estimate a fine pose by transforming the learned 3D point cloud to match a test image.

\subsection{Parametrizing Pose Transformations}\label{sec:pose_sampling}

In order to align our rendered image to the unknown pose of a query image, we need to estimate its camera pose.
Instead of transforming the camera pose directly, we keep the camera fixed to its coarse estimate and transform the entire scene, i.e. the entire 3D point cloud.  This simulates camera movement, allowing us to efficiently align the image rendered by \methodreg{} to the query image.

Concretely, a transformation $T \in SE(3)$ is applied to the position of every Gaussian, as well as to their intrinsic rotation $R$.
All possible transformations from the Euclidean group $SE(3)$ may be modeled using a screw axis $\mathcal{S} = (\omega, v) \in \mathbb{R}^6$ along which to rotate and a distance $\theta \in \mathbb{R}$ to translate along said axis~\cite{ModernRobotics}.
According to Rodrigues' formula for rotations~\cite{ModernRobotics, multi-view-geom}, sampling from the group of 3D rotations $SO(3)$ is done by sampling such a scalar $\theta$ and a unit vector $\omega \in \mathbb{R}^3$ and calculating
\begin{equation}
    Rot(\omega, \theta) = e^{\left[\omega\right]\theta} = I + \sin\theta\left[\omega\right] + (1 - \cos\theta)\left[\omega\right]^2,
\end{equation}
where $\left[\omega\right]$ is the skew-symmetric $3\times3$ matrix representation of $\omega$.
The full transformation $T \in SE(3)$ is then parametrized as

\begin{equation}
    T = e^{\left[S\right]\theta} = \begin{bmatrix}
                                    e^{\left[\omega\right]\theta} & G\left(\theta\right)v \\
                                    0 & 1 
                                \end{bmatrix} \text{, where}
\end{equation}
\begin{equation}
    G\left(\theta\right) = I\theta + (1 - \cos\theta) \left[\omega\right] + (\theta - \sin\theta)\left[\omega\right]^2,
\end{equation}  
which can be directly applied to our 3D point cloud~\cite{ModernRobotics}.
Thereby, any gradients flowing to the 3D point cloud of Gaussians can also be propagated back to the parametrization of our transformation $T$, which is given by the parameters $\omega, v \in \mathbb{R}^3$ and $\theta \in \mathbb{R}$.

\subsection{Anomaly Detection with \methodreg{}}

Since \glspl{nerf} are too resource-intensive for practical use~\cite{NERF, madsim}, \methodreg{} uses \gls{3dgs} to encode the multi-view images of the objects within a 3D point cloud of Gaussians.
Apart from following the standard representations proposed by Kerbl \etal~\cite{3DGS}, the spherical harmonics, which represent the colors, are restricted to a degree of zero.
This restriction still suffices for good image rendering while saving compute power and keeping the color invariant to the rotations that are applied during pose estimation.

\begin{table*}[ht]
\centering
\resizebox{\linewidth}{!}{
\begin{tabular}{l|cc|cc|ccc|cc|cc|cc}
\multicolumn{1}{l}{} & \multicolumn{11}{c}{\leftarrowfill 2D Image-based without Poses \rightarrowfill} & \multicolumn{2}{c}{\leftarrowfill 3D Object-based\rightarrowfill} \\
\toprule
  &  \multicolumn{2}{c|}{{\textbf{Feature Emb.}}} &  \multicolumn{2}{c|}{{\textbf{Memory Banks}}}          & \multicolumn{3}{c|}{{\textbf{Student-Teacher}}}  & \multicolumn{2}{c|}{{\textbf{Normal. Flow}}} & \multicolumn{2}{c|}{{\textbf{Synthetic}}} & \multicolumn{2}{c}{{\textbf{View Synthesis}}}  \\   
    & PaDiM  &   Mahal.  &  CFA & PatchCore &  RD4AD & AST & STFPM &  CFlow      & CS-Flow     & SimpleNet  & DR\AE M   & OmniAD   &  \methodreg{}   \\
  \multirow{-3}{*}{Category} & \cite{AD:padim} & \cite{AD:maha_gde}  & \cite{AD:CFA} & \cite{AD:patchcore} & \cite{AD:RD4AD} &  \cite{AD:AST} &\cite{AD:STFPM} & \cite{AD:cflow-ad} & \cite{AD:csflow} & \cite{AD:simplenet} & \cite{AD:draem} & \cite{madsim} & \textbf{(ours)} \\\midrule 
    Gorilla    & 46.9 & 32.4 & 41.8 & 66.8 & 51.9 & 28.3 & 65.3 & 69.2 & 41.9 & 40.9 & 58.9 & \textbf{93.6} & \underline{91.7} $\pm$ 1.1  \\
    Unicorn    & 81.0 & 86.4 & 85.6 & 92.4 & 67.7 & 87.0 & 79.6 & 82.3 & 85.5 & 88.7 & 70.4 & \underline{94.0} & \textbf{97.9} $\pm$ 1.1  \\
    Mallard    & 14.8 & \underline{89.1} & 36.6 & 59.3 & 54.4 & 48.6 & 42.2 & 74.9 & 36.8 & 43.4 & 34.5 & 84.7 & \textbf{97.4} $\pm$ 0.5  \\
    Turtle     & 54.7 & 21.2 & 58.3 & 87.0 & 82.6 & 28.1 & 64.4 & 51.0 & 56.0 & 62.9 & 18.4 & \underline{95.6} & \textbf{97.2} $\pm$ 0.7  \\
    Whale      & 75.7 & 35.0 & 77.7 & \underline{86.0} & 64.1 & 26.6 & 64.1 & 57.0 & 47.4 & 78.5 & 65.8 & 82.5 & \textbf{95.4} $\pm$ 3.0  \\
    Bird       & 55.6 & 79.7 & 78.4 & 82.9 & 59.8 & 87.1 & 52.4 & 75.6 & 84.9 & 76.3 & 69.1 & \underline{92.4} & \textbf{94.0} $\pm$ 1.2  \\
    Owl        & 75.8 & 67.0 & 74.0 & 72.9 & 69.0 & 81.8 & 72.7 & 76.5 & 71.9 & 74.1 & 67.2 & \textbf{88.2} & \underline{86.8} $\pm$ 0.9  \\
    Sabertooth & 65.4 & 75.8 & 64.2 & 76.6 & 69.9 & 82.2 & 56.0 & 71.3 & 73.9 & 69.6 & 68.6 & \textbf{95.7} & \underline{95.2} $\pm$ 1.5  \\
    Swan       & 59.7 & 70.5 & 66.7 & 75.2 & 60.5 & 83.0 & 53.6 & 67.4 & 65.9 & 66.2 & 59.7 & \underline{86.5} & \textbf{93.0} $\pm$ 0.7  \\
    Sheep      & 69.1 & 82.0 & 86.5 & 89.4 & 77.7 & \underline{96.0} & 56.5 & 80.9 & 84.5 & 81.1 & 59.5 & 90.1 & \textbf{96.7} $\pm$ 0.1  \\
    Pig        & 50.4 & 61.5 & 66.7 & 85.7 & 59.0 & 72.9 & 50.6 & 72.1 & 68.2 & 65.9 & 64.4 & \underline{88.3} & \textbf{96.1} $\pm$ 1.9  \\
    Zalika     & 39.0 & 63.0 & 52.1 & 68.2 & 49.1 & 68.1 & 53.7 & 66.9 & 47.9 & 62.4 & 51.7 & \underline{88.2} & \textbf{89.9} $\pm$ 0.7  \\
    Phoenix    & 61.1 & 63.7 & 65.9 & 71.4 & 62.0 & 74.8 & 56.7 & 64.4 & 63.7 & 62.5 & 53.1 & \underline{82.3} & \textbf{84.2} $\pm$ 0.3  \\
    Elephant   & 48.9 & 70.3 & 71.7 & 78.6 & 62.3 & 86.5 & 61.7 & 70.1 & 71.0 & 71.7 & 62.5 & \underline{92.5} & \textbf{94.7} $\pm$ 0.9  \\
    Parrot     & 53.9 & 64.1 & 69.8 & 78.0 & 47.8 & 76.0 & 61.1 & 67.9 & 59.4 & 66.2 & 62.3 & \textbf{97.0} & \underline{96.1} $\pm$ 1.1  \\
    Cat        & 53.6 & 54.2 & 68.2 & 78.7 & 54.6 & 70.2 & 52.2 & 65.8 & 61.0 & 64.5 & 61.3 & \textbf{84.9} & \underline{82.4} $\pm$ 1.3  \\
    Scorpion   & 80.4 & 78.4 & 91.4 & 82.1 & 69.7 & 90.1 & 68.9 & 79.5 & 76.9 & 80.5 & 83.7 & \underline{91.5} & \textbf{99.2} $\pm$ 0.1  \\
    Obesobeso  & 68.8 & 69.7 & 80.6 & 89.5 & 77.6 & 86.3 & 60.8 & 80.0 & 77.4 & 81.0 & 73.9 & \textbf{97.1} & \underline{95.7} $\pm$ 0.7  \\
    Bear       & 65.5 & 74.4 & 78.7 & 84.2 & 67.5 & 87.4 & 60.7 & 81.4 & 75.3 & 79.4 & 76.1 & \underline{98.8}&\textbf{98.9} $\pm$ 0.2  \\
    Puppy      & 42.9 & 63.3 & 53.7 & 65.6 & 60.3 & 67.9 & 56.7 & 71.4 & 64.1 & 61.0 & 57.4 & \underline{93.5} & \textbf{96.1} $\pm$ 0.9  \\\midrule
    mean         & 58.2 & 65.1 & 68.4 & 78.5 & 63.4 & 71.4 & 59.5 & 71.3 & 65.7 & 68.8 & 60.9 & \underline{90.9} & \textbf{93.9} $\pm$ 0.2  \\
\end{tabular}
}
\caption{AUROC $\left(\uparrow\right)$ for image-level anomaly detection performance on MAD. We run our approach $n=5$ times and report the standard deviation and mark the best result per class in \textbf{bold} and the runner-up \underline{underlined}.}
\label{tab:image_aurocs}
\vspace*{-\baselineskip}
\end{table*}

With the ability to synthesize high-quality views of arbitrary poses using our 3D point cloud, the pose of a test query image $I_q$ still remains unknown.
To recover an initial coarse pose, \methodreg{} sticks to OmniAD's proposed method using \gls{loftr}, which also lets us better estimate the impact of replacing both \glspl{nerf} and iNeRF with our \gls{3dgs}-based framework.

With an initial coarse pose fixed, transformations $T$ from the group of 3D motions are sampled as described in~\cref{sec:pose_sampling}, and applied to each Gaussian in the 3D point cloud.
In each time step $i$, using the transformation $T_i$, a new sample $I_i$ is generated, which may be compared to the query image $I_q$.
For this, we again stick to the original \gls{3dgs} framework and use the mean absolute error $\mathcal{L}_1$ and \gls{ssim}~\cite{SSIM} to optimize the pose with the loss
\begin{equation}\label{eq:loss}
    \mathcal{L} = (1 - \lambda)\mathcal{L}_1 + \lambda\left(1 - \mathcal{L}_{SSIM}\right).
\end{equation}
In the standard \gls{3dgs} setting, the loss is propagated solely back to the 3D point cloud.
By modifying the cloud using our transformations $T$, the gradients instead flow back to the pose transformation sampler, making the entire pose estimation process differentiable. Contrary to iNeRF, no region-of-interest sampling is needed and the entire scene can be used for optimization.

After $k$ steps of pose estimation, a rendering $I_k$, which has the same object pose as $I_q$, but without any anomalies, is generated. Lastly, similar to OmniAD, an anomaly score map is constructed by pixel-wise comparison of extracted pre-trained features on both $I_q$ and $I_k$, again using the mean squared error.

\section{Experiments}\label{sec:experiments}

We evaluate the anomaly detection, segmentation, and pose estimation performance of \methodreg{} empirically on the recently proposed pose-agnostic anomaly detection 
benchmark~\cite{madsim}.

\subsection{Evaluation Protocol}

As is standard in anomaly detection settings, only anomaly-free data is available at training time. For evaluation, both defective and normal samples are given.
The anomaly detection performance is measured by producing an anomaly score for each sample, separating normal data and anomalies. We calculate the area under the ROC curve (AUROC) using this score.
Similarly, as ground truth masks of the anomaly positions are available, the pixel-wise performance can be evaluated by calculating the AUROC across all pixels.
In practice, however, all kinds of anomalies should be detected with equal importance regardless of their size. To treat smaller anomalies with the same weight as larger ones, AUPRO is a well-established metric~\cite{MVTecAD}. It assigns the same weight to each of the connected components within the anomaly masks, making it more difficult to reach high scores and we report it for all models reproduced by us.

\begin{table*}
    \resizebox{\linewidth}{!}{
    \begin{tabular}{l|c|cc|ccc|cc|cc|cc}
    \toprule
  &  \multicolumn{1}{c|}{{\textbf{Feat.Emb.}}} &  \multicolumn{2}{c|}{{\textbf{Memory banks}}}          & \multicolumn{3}{c|}{{\textbf{Student-Teacher}}}  & \multicolumn{2}{c|}{{\textbf{Normal. Flow}}} & \multicolumn{2}{c|}{{\textbf{Synthetic}}} & \multicolumn{2}{c}{{\textbf{View Synthesis}}}  \\   
   & PaDiM     &  CFA & PatchCore & RD4AD & AST & STFPM &  CFlow      & CS-Flow         & SimpleNet  & DR\AE M  & OmniAD   &   \methodreg{}  \\
  \multirow{-3}{*}{Metric} & \cite{AD:padim} & \cite{AD:CFA} & \cite{AD:patchcore} & \cite{AD:RD4AD} &  \cite{AD:AST} & \cite{AD:STFPM} & \cite{AD:cflow-ad} & \cite{AD:csflow} & \cite{AD:simplenet} &  \cite{AD:draem} & \cite{madsim} & \textbf{(ours)} \\\midrule
    AUROC $(\uparrow)$  & 90.7	& 89.4 & 74.9 & 91.3 & 	68.1	& 89.3	& 90.8	& 71.7	& 89.7	& 58.0	& \underline{98.4}	& \textbf{99.5}	$\pm$ 0.01 \\
    AUPRO $(\uparrow)$ & 77.0 & - & 83.7 & 79.2 & 68.1 & - & - & 27.1 & 77.4 & - & \underline{86.6} & \textbf{95.8} $\pm$ 0.03 
 \\
    \end{tabular}
    }
    \caption{Results for anomaly segmentation averaged across all categories in MAD. We run our approach $n = 5$ times and mark the best result in \textbf{bold} and the runner-up \underline{underlined}. Comprehensive tables are in the supplementary in~\cref{sec:supplement_AD}.}
    \label{tab:pixel-wise}
    \vspace*{-\baselineskip}
\end{table*}

\subsection{Data Sets}\label{sec:datasets}

We test our method on the Multi-pose Anomaly Detection (MAD) data set proposed by Zhou \etal~\cite{madsim}, which introduced the challenge of performing pose-agnostic detection in images. To this end, the MAD-Sim data set contains images of 20 different kinds of LEGO\textregistered{} toys of different complexities in terms of shape and texture. All of these were generated synthetically using Blender. At training time, the camera poses for all images are given, whereas they are not available at test time. A few examples of anomalies can be found in~\cref{fig:teaser}. This benchmark is motivated by all major anomaly detection data sets featuring tightly aligned poses, which may interfere with detecting partially occluded anomalies on more complex-shaped objects.

To verify our pose estimation more thoroughly, we also use the 360° synthetic scenes provided by NeRF~\cite{NERF}, which are already divided into train and test splits. These also contain synthetically generated multi-view images but are not constrained to simple LEGO\textregistered{} toys, as they also include more complex objects with finer details.

In our experiments, we find that for both data sets, the poses in the training split are uniformly distributed in a sphere around the object. For MAD, the test poses are sampled randomly within that sphere without any well-defined distribution, signifying the need for the pose estimation to be able to generalize to previously unseen poses. The test poses in the NeRF data set follow the same uniform distribution as the train set and instead describe one circular orbit around the objects.

\subsection{Implementation}\label{sec:impl}

We base our implementation on the reference given by OmniAD\footnote{\url{https://github.com/EricLee0224/PAD}}~\cite{madsim}.
Implementations for Gaussian Splatting are taken from the original project by Kerbl \etal\footnote{\url{https://github.com/graphdeco-inria/gaussian-splatting}}\cite{3DGS}, which we modify to allow for the insertion of pose transformations into the differentiable rendering process. 

As for the modules replicated from OmniAD, coarse pose estimation is done using a pre-trained \gls{loftr}~\cite{LOFTR} with a ResNet~\cite{ResNet} backbone, while the image feature matching is done using a pre-trained EfficientNet-b4~\cite{EfficientNet}.
Both are pre-trained on ImageNet~\cite{imagenet}. All images are also resized to $400\times400$ pixels before processing.

We train our 3D Gaussian model for $30,000$ iterations while keeping everything to the author-given hyperparameters, except for the spherical harmonics~\cite{3DGS}.
The pose estimation is optimized using the Adam optimizer~\cite{Adam_opt} with a learning rate of $0.001$ and momentum parameters $\beta_1 = 0.9$ and $\beta_2=0.999$.
For the loss in~\cref{eq:loss}, we use $\lambda = 0.2$.
The pose estimation iterates for $k = 175$ steps, which we find to yield good estimations while still allowing for quick inference times.

\begin{figure}
    \centering
    \includegraphics[width=0.85\linewidth]{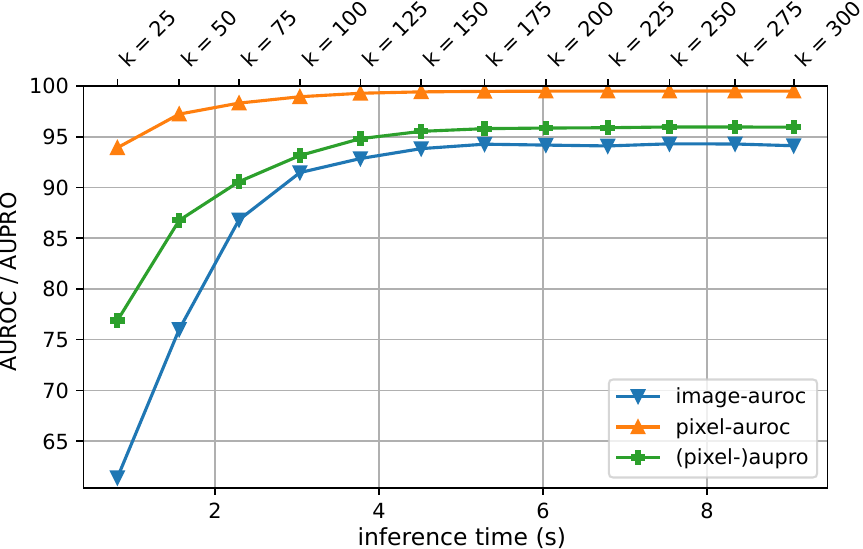}
    \caption{Influence on all detection metrics and inference speed, when changing the number of pose estimation steps $k$ from $25$ to $300$, with the performance saturating around $k = 175$. Since OmniAD has inference times magnitudes larger than \methodreg{}, we do not include it in this experiment.}
    \label{fig:ablation_k}
    \vspace*{-\baselineskip}
\end{figure}

\subsection{Anomaly Detection Results}\label{sec:results}

Detailed results for image-wise anomaly detection on \gls{mad} are given in~\cref{tab:image_aurocs}. We have taken numbers for competitors from Zhou \etal~\cite{madsim} and also reproduced additional state-of-the-art models ourselves.
Our method confidently reaches a new state-of-the-art performance of $93.9\%$, which is three points higher than the next best competitor OmniAD.
It should also be noted, that all methods that do not incorporate any pose information, i.e. the ones not synthesis-based, lack behind heavily in performance, with PatchCore~\cite{AD:patchcore} reaching the highest AUROC of only $78.5\%$. 
We observe, that \methodreg{} especially outperforms OmniAD, whenever the pose estimation is more accurate.
This intuition will be verified quantitatively in~\cref{sec:4_poseest}.

We furthermore also evaluate the anomaly segmentation performance by looking at the pixel-wise AUROC and AUPRO in~\cref{tab:pixel-wise}. Again \methodreg{} clearly outperforms all competitors with a pixel-wise AUROC of $99.5\%$ and an AUPRO of $95.8$. Some qualitative intuition of the superior performance of \methodreg{} will later be visualized in~\cref{fig:subset_anomaly_maps}.

\subsection{Computational Impact}\label{sec:compute}

One of the major shortcomings of our main competitor OmniAD is its high training and inference times, which we overcome with our approach.
Therefore, we evaluated training and inference times across our experiments in~\cref{tab:speed}.
For a fair comparison, all experiments are executed on the same machines, utilizing a single NVIDIA RTX 3090. 

Since both our and the OmniAD pipeline perform the same coarse pose estimation, we only measure the inference times caused by the iterative fine pose estimation and the subsequent feature matching.
We closely match the computational requirements reported by Zhou \etal for OmniAD~\cite{madsim}.
As our approach takes far fewer optimization steps both when training the cloud of Gaussians and when fine-tuning the estimated pose, we can confidently beat their computational requirements.
In total, it results in roughly $\textbf{55}$ times faster training and $\textbf{13}$ times faster inference when using our method.
Furthermore, our \gls{3dgs}-based training always converged to good solutions, while \gls{nerf} struggled to converge in roughly $15\%$ of training runs due to poor ray sampling.
\methodreg's robustness and speed advantage allow for both quick prototyping and on-the-fly anomaly detection in real production scenarios.

Furthermore, we run an ablation on the number of steps used for pose estimation $k$ in~\cref{fig:ablation_k}.
Starting with $k=25$, we observe an image-wise AUROC of $61\%$ and inference times of roughly one second.
While inference time linearly increases with the number of steps, the detection performance improves drastically and starts to saturate after $k = 175$, which we suggest to be a good compromise between quick inference and precise detection.
Higher values of $k$ do not substantially improve the performance, as the pose is already very precisely estimated.

\begin{table}[t]
    \centering
    \begin{tabular}{c|cc}
    \toprule
      &   training $\downarrow$ & inference $\downarrow$ \\  
      \multirow{-2}{*}{Method}  & $(\text{h:min:sec})_{\pm (\text{min}:\text{sec})}$ & $\text{(min:sec})_{\pm\text{(sec)}}$ \\\midrule
       OmniAD & $\text{04:33:43}_{\pm 04:52}$ & $\text{01:06}_{\pm 00}$\\
       \methodreg{} & $\textbf{00:04:54}_{\pm 00:18}$ & $\textbf{00:05}_{\pm 00}$\\
    \end{tabular}
    \caption{Both training and inference speed measured on the MAD data set for OmniAD~\cite{madsim} and our approach. One standard deviation is given across the 20 different classes. Best results in \textbf{bold}.}
    \label{tab:speed}
\end{table}

\subsection{Pose Estimation Results}\label{sec:4_poseest}

Despite the focus of our work being on anomaly detection, we evaluate the pose estimation process more thoroughly and compare the performance of our method and iNeRF.

As commonly done in pose estimation literature, we compare the location of the object whose pose we estimate (here the camera, as we only have access to ground truth camera poses) to the ground truth position using the Euclidean distance~\cite{6dof_add}. 

For the estimated rotation matrices, we first convert all rotation matrices into their representation as unit quaternions.
For two quaternions $q_1$ and $q_2$, we can then quantify the difference between two rotations, by measuring the rotation needed to go from $q_1$ to $q_2$, resulting in the formula
\begin{equation}\label{eq:rot_metric}
    \Phi(q_1, q_2) = 2\arccos\left(\left|q_1 \cdot q_2\right|\right),
\end{equation}
with $\cdot$ denoting the dot product of two vectors.
$\Phi$ defines a metric on the group of 3D rotations $SO(3)$ and takes values in the ranges of $[0, \pi]$ with lower scores denoting a higher similarity in rotation~\cite{rotation_metrics}.

\begin{figure}[h]
    \centering
    \includegraphics[width=0.8\linewidth]{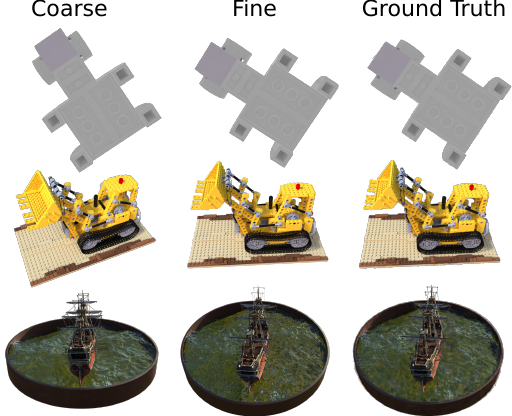}
    \caption{Examples of pose estimation using \methodreg{}. Starting from OmniAD's coarse pose~\cite{madsim}, we refine it to match the ground truth. Examples are from MAD and the NeRF synthetic data.}
    \label{fig:pose_est_viz}
\end{figure}

A few estimated poses are visualized in~\cref{fig:pose_est_viz}. As no true ground truth poses are available for the MAD data set, we randomly split the training data set into a custom training and testing split to validate our methods.
Quantitative results of the experiments are presented in~\cref{tab:pose_est}, with the full table in the supplementary in~\cref{sec:supplement_pose_estimation}.

\begin{table}[ht]
    \begin{tabular}{c|p{1.2cm}p{1.2cm}|p{1cm}p{1cm}}
        \toprule
         &  \multicolumn{2}{c|}{Transl. Err. (total) $\downarrow$} &   \multicolumn{2}{c}{Rot. Err. (rad) $\downarrow$}\\\cmidrule{2-5}
       \multirow{-2}{*}{Method} & MAD & Synth & MAD & Synth \\\midrule
       Coarse~\cite{madsim} & 0.803 & 0.624 & 0.163 & 0.084\\
       iNeRF~\cite{iNERF} & 0.179 & 0.055 & 0.056 & 0.007\\
       \methodreg{} & \textbf{0.094} & \textbf{0.021} & \textbf{0.040} & \textbf{0.003}\\
    \end{tabular}
    \caption{Pose Estimation Error measured as the Euclidean distance for Translation and using $\Phi$ from~\cref{eq:rot_metric} for Rotation. Results are averaged across the MAD data set~\cite{madsim} and \gls{nerf} synthetic scenes~\cite{NERF}. Best results in \textbf{bold}.}
    \label{tab:pose_est}
\end{table}

We observe, that the coarse pose estimation step of OmniAD is already able to achieve good baseline scores.
Still, both iNeRF and \methodreg{} can vastly outperform the rough estimation, with us performing the best out of all approaches, despite using fewer optimization steps and very short training times.
According to the full results in the supplements, we outperform iNeRF by several magnitudes in most of the categories, with closer margins only occurring in a few cases.
We argue that this is caused by us being able to optimize using the entire image and full 3D point cloud, while iNeRF is restricted to sampling the most important rays, which misses out on some details. 
\begin{figure}[ht!]
    \centering
    \includegraphics[width=\linewidth]{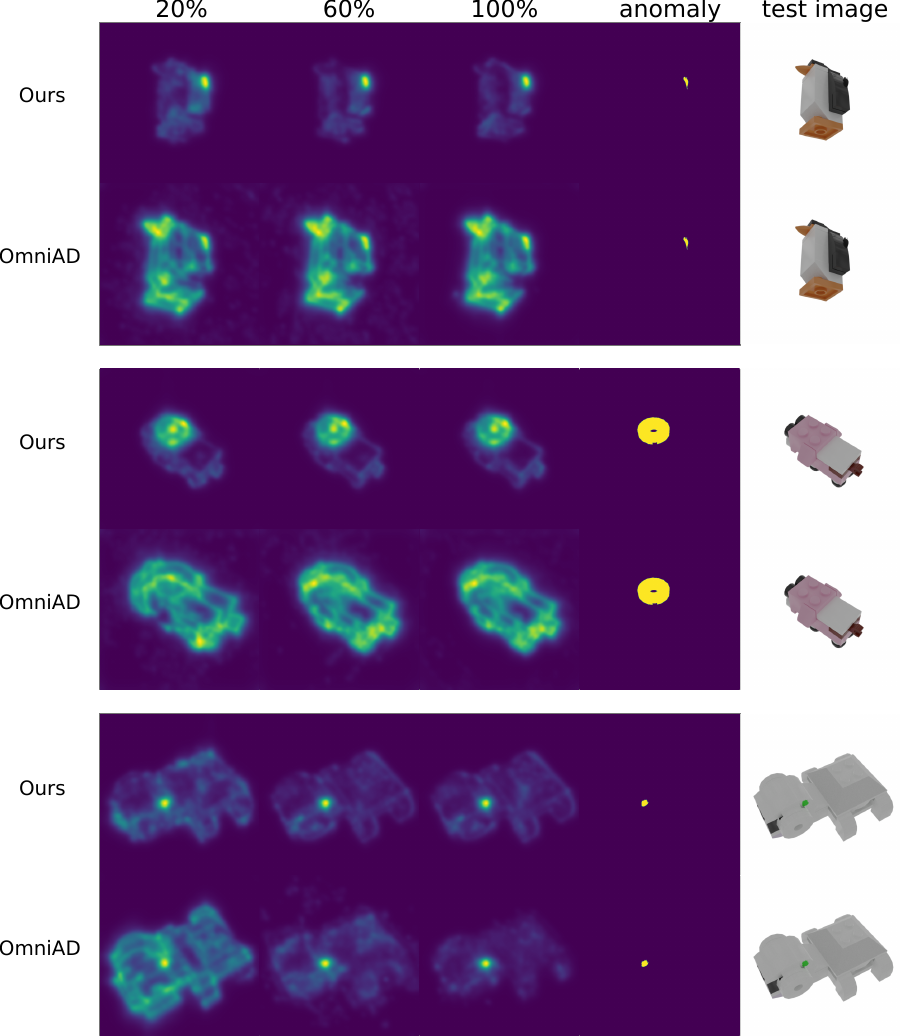}
    \caption{Quantitative comparison of performance for both OmniAD when using between $20\%$ and $100\%$ of the available training data. We show the anomaly maps achieved by feature matching for both methods. Best viewed in color.}
    \label{fig:subset_anomaly_maps}
\end{figure}

The quantitative comparison for the synthetic \gls{nerf} scenes from~\cref{tab:pose_est}, again have \methodreg{} performing better.
For both the \gls{nerf}'s uniformly sampled test scenes and the randomly sampled ones from MAD, \methodreg{} achieves a stronger pose estimation performance.
Therefore, it is not only able to better capture the 3D geometry of the scene for known camera poses but also its capability to generalize to previously unseen views beats that of iNeRF.
This superiority in pose estimation also shows up in \methodreg{} beating out OmniAD at the anomaly detection tasks, since the rendered images are aligned more accurately to the test samples. 

\subsection{Results for Sparse-View Data}\label{sec:4_sparse}

Each category in the \gls{mad} data set contains $210$ different views for training.
In practice, multi-view data can not be guaranteed to be this dense.
Thus, we sample subsets of the original training splits, to simulate sparse-view settings.

\begin{table}[ht!]
\begin{tabular}{c|cc|cc}
\toprule
\multicolumn{1}{c|}{\%} & \multicolumn{2}{c|}{Detection} & \multicolumn{2}{c}{Segmentation} \\\cmidrule{2-5}
\multicolumn{1}{c|}{views} & OmniAD & \methodreg{} & OmniAD & \methodreg{} \\
\midrule
20 & 67.2 & \textbf{77.6} & 73.7 & \textbf{86.5}  \\
40 & 79.1 & \textbf{89.3} & 81.5 & \textbf{93.3}  \\
60 & 82.2 & \textbf{92.2} & 85.0 & \textbf{95.0} \\
80 & 86.8 & \textbf{93.1}  & 86.8 & \textbf{95.5}\\
100 & 90.9 & \textbf{93.9} & 86.6 & \textbf{95.8}\\
\end{tabular}
\caption{Image-wise Anomaly Detection performance for the sparse-view setting measured as AUROC ($\uparrow$) and AUPRO ($\uparrow$) respectively on \gls{mad}. Best results in \textbf{bold}.}
\label{tab:subset_imagewise}
\vspace*{-\baselineskip}
\end{table}

Some selected visualizations of the anomaly maps for training with $20, 60,$ and $100\%$ of the data are shown in~\cref{fig:subset_anomaly_maps}. 
Here, our intuition is confirmed, that better pose estimation leads to more precise anomaly maps.
OmniAD struggles most when the coarse pose is too far away, or when large parts of the object are missing.
Even when OmniAD is able to fit a good pose, artifacts and surrounding rendering mistakes are more frequent than with \methodreg{}.

The quantitative results for image-wise anomaly detection and segmentation under a sparse-view setting can be found in~\cref{tab:subset_imagewise}, with the full table in the supplement in~\cref{sec:supplement_sparse}. We compare \methodreg{} to the other top competitor OmniAD, despite the high computational costs~\cite{madsim}.

\methodreg{} decisively beats OmniAD for every step of view-sparsification, by margins of around ten points in the $20$, $40$ and $60\%$ settings for both the image-wise AUROC and AUPRO. For lower levels of sparsity, \methodreg{} still outperforms OmniAD decisively. 
Again, we can optimize using the full views, while \gls{nerf} resorts to sampling rays and is known to struggle in sparse-view settings~\cite{regnerf}.

\section{Conclusion}\label{sec:conclusion}

In this paper, we presented a novel pose-agnostic method for anomaly detection.
Given multi-view images, we represent the object as a 3D point cloud of Gaussians,
which is used for pose estimation, and to find anomalies in images without prior pose information.
Our method beats all competitors at the detection task, while still being magnitudes faster for both training and inference time, making it more suited for deployment in production environments.

We would like to dedicate future work towards improving the coarse pose estimation and image feature comparison.
Applying our findings to adjacent fields, such as pose estimation in humans~\cite{nf_humanpose, human_pose_36m}, strikes us as a promising next direction.
Closing the gap between synthetic and real-world data would also require more work.
Lastly, we want to investigate ways to include the three-dimensional point cloud information in existing two-dimensional approaches.

\vspace*{-0.5\baselineskip}
\small{\paragraph{Acknowledgements.} This work was supported by the Federal Ministry of Education and Research (BMBF), Germany under the AI service center KISSKI (grant no. 01IS22093C), the Lower Saxony Ministry of Science and Culture (MWK) through the zukunft.niedersachsen program of the Volkswagen Foundation and the Deutsche Forschungsgemeinschaft (DFG) under Germany’s Excellence Strategy within the Cluster of Excellence PhoenixD (EXC 2122).}

{
    \small
    \bibliographystyle{ieeenat_fullname}
    \bibliography{main}

\begin{thebibliography}{47}
\providecommand{\natexlab}[1]{#1}
\providecommand{\url}[1]{\texttt{#1}}
\expandafter\ifx\csname urlstyle\endcsname\relax
  \providecommand{\doi}[1]{doi: #1}\else
  \providecommand{\doi}{doi: \begingroup \urlstyle{rm}\Url}\fi

\bibitem[Barron et~al.(2022)Barron, Mildenhall, Verbin, Srinivasan, and Hedman]{MipNeRF360}
Jonathan~T. Barron, Ben Mildenhall, Dor Verbin, Pratul~P. Srinivasan, and Peter Hedman.
\newblock Mip-nerf 360: Unbounded anti-aliased neural radiance fields.
\newblock In \emph{{IEEE/CVF} Conference on Computer Vision and Pattern Recognition, {CVPR} 2022, New Orleans, LA, USA, June 18-24, 2022}, pages 5460--5469. {IEEE}, 2022.

\bibitem[Bergmann et~al.(2019)Bergmann, Fauser, Sattlegger, and Steger]{MVTecAD}
Paul Bergmann, Michael Fauser, David Sattlegger, and Carsten Steger.
\newblock Mvtec ad — a comprehensive real-world dataset for unsupervised anomaly detection.
\newblock In \emph{2019 IEEE/CVF Conference on Computer Vision and Pattern Recognition (CVPR)}, pages 9584--9592, 2019.

\bibitem[Bergmann et~al.(2020)Bergmann, Fauser, Sattlegger, and Steger]{AD:UninfST}
Paul Bergmann, Michael Fauser, David Sattlegger, and Carsten Steger.
\newblock Uninformed students: Student-teacher anomaly detection with discriminative latent embeddings.
\newblock In \emph{2020 {IEEE/CVF} Conference on Computer Vision and Pattern Recognition, {CVPR} 2020, Seattle, WA, USA, June 13-19, 2020}, pages 4182--4191. Computer Vision Foundation / {IEEE}, 2020.

\bibitem[Bergmann et~al.(2022{\natexlab{a}})Bergmann, Batzner, Fauser, Sattlegger, and Steger]{Mvtecloco}
Paul Bergmann, Kilian Batzner, Michael Fauser, David Sattlegger, and Carsten Steger.
\newblock Beyond dents and scratches: Logical constraints in unsupervised anomaly detection and localization.
\newblock \emph{Int. J. Comput. Vis.}, 130\penalty0 (4):\penalty0 947--969, 2022{\natexlab{a}}.

\bibitem[Bergmann et~al.(2022{\natexlab{b}})Bergmann, Jin, Sattlegger, and Steger]{MVTEC3D}
Paul Bergmann, Xin Jin, David Sattlegger, and Carsten Steger.
\newblock The mvtec 3d-ad dataset for unsupervised 3d anomaly detection and localization.
\newblock In \emph{Proceedings of the 17th International Joint Conference on Computer Vision, Imaging and Computer Graphics Theory and Applications, {VISIGRAPP} 2022, Volume 5: VISAPP, Online Streaming, February 6-8, 2022}, pages 202--213. {SCITEPRESS}, 2022{\natexlab{b}}.

\bibitem[Cohen and Hoshen(2020)]{AD:spade}
Niv Cohen and Yedid Hoshen.
\newblock Sub-image anomaly detection with deep pyramid correspondences.
\newblock \emph{CoRR}, abs/2005.02357, 2020.

\bibitem[Defard et~al.(2020)Defard, Setkov, Loesch, and Audigier]{AD:padim}
Thomas Defard, Aleksandr Setkov, Angelique Loesch, and Romaric Audigier.
\newblock Padim: {A} patch distribution modeling framework for anomaly detection and localization.
\newblock In \emph{Pattern Recognition. {ICPR} International Workshops and Challenges - Virtual Event, January 10-15, 2021, Proceedings, Part {IV}}, pages 475--489. Springer, 2020.

\bibitem[Deng and Li(2022)]{AD:RD4AD}
Hanqiu Deng and Xingyu Li.
\newblock Anomaly detection via reverse distillation from one-class embedding.
\newblock In \emph{Proceedings of the IEEE/CVF Conference on Computer Vision and Pattern Recognition (CVPR)}, pages 9737--9746, 2022.

\bibitem[Deng et~al.(2009)Deng, Dong, Socher, Li, Li, and Fei-Fei]{imagenet}
Jia Deng, Wei Dong, Richard Socher, Li-Jia Li, Kai Li, and Li Fei-Fei.
\newblock Imagenet: A large-scale hierarchical image database.
\newblock In \emph{2009 IEEE conference on computer vision and pattern recognition}, pages 248--255. Ieee, 2009.

\bibitem[Gudovskiy et~al.(2022)Gudovskiy, Ishizaka, and Kozuka]{AD:cflow-ad}
Denis~A. Gudovskiy, Shun Ishizaka, and Kazuki Kozuka.
\newblock {CFLOW-AD:} real-time unsupervised anomaly detection with localization via conditional normalizing flows.
\newblock In \emph{{IEEE/CVF} Winter Conference on Applications of Computer Vision, {WACV} 2022, Waikoloa, HI, USA, January 3-8, 2022}, pages 1819--1828. {IEEE}, 2022.

\bibitem[Harltey and Zisserman(2006)]{multi-view-geom}
Andrew Harltey and Andrew Zisserman.
\newblock \emph{Multiple view geometry in computer vision {(2.} ed.)}.
\newblock Cambridge University Press, 2006.

\bibitem[He et~al.(2016)He, Zhang, Ren, and Sun]{ResNet}
Kaiming He, Xiangyu Zhang, Shaoqing Ren, and Jian Sun.
\newblock Deep residual learning for image recognition.
\newblock In \emph{2016 {IEEE} Conference on Computer Vision and Pattern Recognition, {CVPR} 2016, Las Vegas, NV, USA, June 27-30, 2016}, pages 770--778. {IEEE} Computer Society, 2016.

\bibitem[Heckler et~al.(2023)Heckler, K{\"{o}}nig, and Bergmann]{HecklerFeatures}
Lars Heckler, Rebecca K{\"{o}}nig, and Paul Bergmann.
\newblock Exploring the importance of pretrained feature extractors for unsupervised anomaly detection and localization.
\newblock In \emph{{IEEE/CVF} Conference on Computer Vision and Pattern Recognition, {CVPR} 2023 - Workshops, Vancouver, BC, Canada, June 17-24, 2023}, pages 2917--2926. {IEEE}, 2023.

\bibitem[Hinterstoisser et~al.(2012)Hinterstoisser, Lepetit, Ilic, Holzer, Bradski, Konolige, and Navab]{6dof_add}
Stefan Hinterstoisser, Vincent Lepetit, Slobodan Ilic, Stefan Holzer, Gary~R. Bradski, Kurt Konolige, and Nassir Navab.
\newblock Model based training, detection and pose estimation of texture-less 3d objects in heavily cluttered scenes.
\newblock In \emph{Computer Vision - {ACCV} 2012 - 11th Asian Conference on Computer Vision, Daejeon, Korea, November 5-9, 2012, Revised Selected Papers, Part {I}}, pages 548--562. Springer, 2012.

\bibitem[Huynh(2009)]{rotation_metrics}
Du~Q. Huynh.
\newblock Metrics for 3d rotations: Comparison and analysis.
\newblock \emph{J. Math. Imaging Vis.}, 35\penalty0 (2):\penalty0 155--164, 2009.

\bibitem[Ionescu et~al.(2014)Ionescu, Papava, Olaru, and Sminchisescu]{human_pose_36m}
Catalin Ionescu, Dragos Papava, Vlad Olaru, and Cristian Sminchisescu.
\newblock Human3.6m: Large scale datasets and predictive methods for 3d human sensing in natural environments.
\newblock \emph{{IEEE} Trans. Pattern Anal. Mach. Intell.}, 36\penalty0 (7):\penalty0 1325--1339, 2014.

\bibitem[Kerbl et~al.(2023)Kerbl, Kopanas, Leimk{\"{u}}hler, and Drettakis]{3DGS}
Bernhard Kerbl, Georgios Kopanas, Thomas Leimk{\"{u}}hler, and George Drettakis.
\newblock 3d gaussian splatting for real-time radiance field rendering.
\newblock \emph{{ACM} Trans. Graph.}, 42\penalty0 (4):\penalty0 139:1--139:14, 2023.

\bibitem[Kingma and Ba(2015)]{Adam_opt}
Diederik~P. Kingma and Jimmy Ba.
\newblock Adam: {A} method for stochastic optimization.
\newblock In \emph{3rd International Conference on Learning Representations, {ICLR} 2015, San Diego, CA, USA, May 7-9, 2015, Conference Track Proceedings}, 2015.

\bibitem[Lee et~al.(2022)Lee, Lee, and Song]{AD:CFA}
Sungwook Lee, Seunghyun Lee, and Byung~Cheol Song.
\newblock {CFA:} coupled-hypersphere-based feature adaptation for target-oriented anomaly localization.
\newblock \emph{{IEEE} Access}, 10:\penalty0 78446--78454, 2022.

\bibitem[Li et~al.(2021)Li, Sohn, Yoon, and Pfister]{AD:cutpaste}
Chun{-}Liang Li, Kihyuk Sohn, Jinsung Yoon, and Tomas Pfister.
\newblock Cutpaste: Self-supervised learning for anomaly detection and localization.
\newblock In \emph{{IEEE} Conference on Computer Vision and Pattern Recognition, {CVPR} 2021, virtual, June 19-25, 2021}, pages 9664--9674. Computer Vision Foundation / {IEEE}, 2021.

\bibitem[Lin et~al.(2021)Lin, Florence, Barron, Rodriguez, Isola, and Lin]{iNERF}
Yen{-}Chen Lin, Pete Florence, Jonathan~T. Barron, Alberto Rodriguez, Phillip Isola, and Tsung{-}Yi Lin.
\newblock inerf: Inverting neural radiance fields for pose estimation.
\newblock In \emph{{IEEE/RSJ} International Conference on Intelligent Robots and Systems, {IROS} 2021, Prague, Czech Republic, September 27 - Oct. 1, 2021}, pages 1323--1330. {IEEE}, 2021.

\bibitem[Liu et~al.(2023)Liu, Zhou, Xu, and Wang]{AD:simplenet}
Zhikang Liu, Yiming Zhou, Yuansheng Xu, and Zilei Wang.
\newblock Simplenet: A simple network for image anomaly detection and localization.
\newblock In \emph{Proceedings of the IEEE/CVF Conference on Computer Vision and Pattern Recognition (CVPR)}, pages 20402--20411, 2023.

\bibitem[Lynch and Park(2017)]{ModernRobotics}
Kevin~M. Lynch and Frank~C. Park.
\newblock \emph{Modern Robotics - Mechanics, Planning, and Control}.
\newblock Cambridge University Press, Cambridge, 2017.

\bibitem[Mildenhall et~al.(2020)Mildenhall, Srinivasan, Tancik, Barron, Ramamoorthi, and Ng]{NERF}
Ben Mildenhall, Pratul~P. Srinivasan, Matthew Tancik, Jonathan~T. Barron, Ravi Ramamoorthi, and Ren Ng.
\newblock Nerf: Representing scenes as neural radiance fields for view synthesis.
\newblock In \emph{Computer Vision - {ECCV} 2020 - 16th European Conference, Glasgow, UK, August 23-28, 2020, Proceedings, Part {I}}, pages 405--421. Springer, 2020.

\bibitem[Niemeyer et~al.(2022)Niemeyer, Barron, Mildenhall, Sajjadi, Geiger, and Radwan]{regnerf}
Michael Niemeyer, Jonathan~T. Barron, Ben Mildenhall, Mehdi S.~M. Sajjadi, Andreas Geiger, and Noha Radwan.
\newblock Regnerf: Regularizing neural radiance fields for view synthesis from sparse inputs.
\newblock In \emph{{IEEE/CVF} Conference on Computer Vision and Pattern Recognition, {CVPR} 2022, New Orleans, LA, USA, June 18-24, 2022}, pages 5470--5480. {IEEE}, 2022.

\bibitem[Perera et~al.(2019)Perera, Nallapati, and Xiang]{AD:OCGAN}
Pramuditha Perera, Ramesh Nallapati, and Bing Xiang.
\newblock {OCGAN:} one-class novelty detection using gans with constrained latent representations.
\newblock In \emph{{IEEE} Conference on Computer Vision and Pattern Recognition, {CVPR} 2019, Long Beach, CA, USA, June 16-20, 2019}, pages 2898--2906. Computer Vision Foundation / {IEEE}, 2019.

\bibitem[Pumarola et~al.(2021)Pumarola, Corona, Pons{-}Moll, and Moreno{-}Noguer]{nerf_dnerf}
Albert Pumarola, Enric Corona, Gerard Pons{-}Moll, and Francesc Moreno{-}Noguer.
\newblock D-nerf: Neural radiance fields for dynamic scenes.
\newblock In \emph{{IEEE} Conference on Computer Vision and Pattern Recognition, {CVPR} 2021, virtual, June 19-25, 2021}, pages 10318--10327. Computer Vision Foundation / {IEEE}, 2021.

\bibitem[Rippel et~al.(2020)Rippel, Mertens, and Merhof]{AD:maha_gde}
Oliver Rippel, Patrick Mertens, and Dorit Merhof.
\newblock Modeling the distribution of normal data in pre-trained deep features for anomaly detection.
\newblock In \emph{25th International Conference on Pattern Recognition, {ICPR} 2020, Virtual Event / Milan, Italy, January 10-15, 2021}, pages 6726--6733. {IEEE}, 2020.

\bibitem[Rosenhahn and Hirche(2024)]{quantum_nf}
Bodo Rosenhahn and Christoph Hirche.
\newblock Quantum normalizing flows for anomaly detection.
\newblock \emph{CoRR}, abs/2402.02866, 2024.

\bibitem[Roth et~al.(2022)Roth, Pemula, Zepeda, Sch{\"{o}}lkopf, Brox, and Gehler]{AD:patchcore}
Karsten Roth, Latha Pemula, Joaquin Zepeda, Bernhard Sch{\"{o}}lkopf, Thomas Brox, and Peter~V. Gehler.
\newblock Towards total recall in industrial anomaly detection.
\newblock In \emph{{IEEE/CVF} Conference on Computer Vision and Pattern Recognition, {CVPR} 2022, New Orleans, LA, USA, June 18-24, 2022}, pages 14298--14308. {IEEE}, 2022.

\bibitem[Rudolph et~al.(2021)Rudolph, Wandt, and Rosenhahn]{AD:differnet}
Marco Rudolph, Bastian Wandt, and Bodo Rosenhahn.
\newblock Same same but differnet: Semi-supervised defect detection with normalizing flows.
\newblock In \emph{{IEEE} Winter Conference on Applications of Computer Vision, {WACV} 2021, Waikoloa, HI, USA, January 3-8, 2021}, pages 1906--1915. {IEEE}, 2021.

\bibitem[Rudolph et~al.(2022)Rudolph, Wehrbein, Rosenhahn, and Wandt]{AD:csflow}
Marco Rudolph, Tom Wehrbein, Bodo Rosenhahn, and Bastian Wandt.
\newblock Fully convolutional cross-scale-flows for image-based defect detection.
\newblock In \emph{{IEEE/CVF} Winter Conference on Applications of Computer Vision, {WACV} 2022, Waikoloa, HI, USA, January 3-8, 2022}, pages 1829--1838. {IEEE}, 2022.

\bibitem[Rudolph et~al.(2023)Rudolph, Wehrbein, Rosenhahn, and Wandt]{AD:AST}
Marco Rudolph, Tom Wehrbein, Bodo Rosenhahn, and Bastian Wandt.
\newblock Asymmetric student-teacher networks for industrial anomaly detection.
\newblock In \emph{2023 IEEE/CVF Winter Conference on Applications of Computer Vision (WACV)}. IEEE, 2023.

\bibitem[Schlegl et~al.(2017)Schlegl, Seeb{\"{o}}ck, Waldstein, Schmidt{-}Erfurth, and Langs]{AD:anogan}
Thomas Schlegl, Philipp Seeb{\"{o}}ck, Sebastian~M. Waldstein, Ursula Schmidt{-}Erfurth, and Georg Langs.
\newblock Unsupervised anomaly detection with generative adversarial networks to guide marker discovery.
\newblock In \emph{Information Processing in Medical Imaging - 25th International Conference, {IPMI} 2017, Boone, NC, USA, June 25-30, 2017, Proceedings}, pages 146--157. Springer, 2017.

\bibitem[Sch{\"{o}}nberger and Frahm(2016)]{SFM:colmap}
Johannes~L. Sch{\"{o}}nberger and Jan{-}Michael Frahm.
\newblock Structure-from-motion revisited.
\newblock In \emph{2016 {IEEE} Conference on Computer Vision and Pattern Recognition, {CVPR} 2016, Las Vegas, NV, USA, June 27-30, 2016}, pages 4104--4113. {IEEE} Computer Society, 2016.

\bibitem[Sch{\"{o}}nberger et~al.(2016)Sch{\"{o}}nberger, Zheng, Frahm, and Pollefeys]{MVS:colmap}
Johannes~L. Sch{\"{o}}nberger, Enliang Zheng, Jan{-}Michael Frahm, and Marc Pollefeys.
\newblock Pixelwise view selection for unstructured multi-view stereo.
\newblock In \emph{Computer Vision - {ECCV} 2016 - 14th European Conference, Amsterdam, The Netherlands, October 11-14, 2016, Proceedings, Part {III}}, pages 501--518. Springer, 2016.

\bibitem[Snavely et~al.(2006)Snavely, Seitz, and Szeliski]{SFM:Snavely06}
Noah Snavely, Steven~M. Seitz, and Richard Szeliski.
\newblock Photo tourism: exploring photo collections in 3d.
\newblock \emph{{ACM} Trans. Graph.}, 25\penalty0 (3):\penalty0 835--846, 2006.

\bibitem[Sun et~al.(2021)Sun, Shen, Wang, Bao, and Zhou]{LOFTR}
Jiaming Sun, Zehong Shen, Yuang Wang, Hujun Bao, and Xiaowei Zhou.
\newblock Loftr: Detector-free local feature matching with transformers.
\newblock In \emph{{IEEE} Conference on Computer Vision and Pattern Recognition, {CVPR} 2021, virtual, June 19-25, 2021}, pages 8922--8931. Computer Vision Foundation / {IEEE}, 2021.

\bibitem[Tan and Le(2019)]{EfficientNet}
Mingxing Tan and Quoc~V. Le.
\newblock Efficientnet: Rethinking model scaling for convolutional neural networks.
\newblock In \emph{Proceedings of the 36th International Conference on Machine Learning, {ICML} 2019, 9-15 June 2019, Long Beach, California, {USA}}, pages 6105--6114. {PMLR}, 2019.

\bibitem[Wang et~al.(2021)Wang, Han, Ding, and Huang]{AD:STFPM}
Guodong Wang, Shumin Han, Errui Ding, and Di Huang.
\newblock Student-teacher feature pyramid matching for anomaly detection.
\newblock In \emph{32nd British Machine Vision Conference 2021, {BMVC} 2021, Online, November 22-25, 2021}, page 306. {BMVA} Press, 2021.

\bibitem[Wang et~al.(2023)Wang, Peng, Zhang, Yi, Wang, and Wang]{AD:HybridFusion}
Yue Wang, Jinlong Peng, Jiangning Zhang, Ran Yi, Yabiao Wang, and Chengjie Wang.
\newblock Multimodal industrial anomaly detection via hybrid fusion.
\newblock In \emph{{IEEE/CVF} Conference on Computer Vision and Pattern Recognition, {CVPR} 2023, Vancouver, BC, Canada, June 17-24, 2023}, pages 8032--8041. {IEEE}, 2023.

\bibitem[Wang et~al.(2004)Wang, Bovik, Sheikh, and Simoncelli]{SSIM}
Zhou Wang, Alan~C. Bovik, Hamid~R. Sheikh, and Eero~P. Simoncelli.
\newblock Image quality assessment: from error visibility to structural similarity.
\newblock \emph{{IEEE} Trans. Image Process.}, 13\penalty0 (4):\penalty0 600--612, 2004.

\bibitem[Wehrbein et~al.(2021)Wehrbein, Rudolph, Rosenhahn, and Wandt]{nf_humanpose}
Tom Wehrbein, Marco Rudolph, Bodo Rosenhahn, and Bastian Wandt.
\newblock Probabilistic monocular 3d human pose estimation with normalizing flows.
\newblock In \emph{2021 {IEEE/CVF} International Conference on Computer Vision, {ICCV} 2021, Montreal, QC, Canada, October 10-17, 2021}, pages 11179--11188. {IEEE}, 2021.

\bibitem[Weng et~al.(2022)Weng, Curless, Srinivasan, Barron, and Kemelmacher-Shlizerman]{nerf_humannerf}
Chung-Yi Weng, Brian Curless, Pratul~P. Srinivasan, Jonathan~T. Barron, and Ira Kemelmacher-Shlizerman.
\newblock Human{N}e{RF}: Free-viewpoint rendering of moving people from monocular video.
\newblock In \emph{Proceedings of the IEEE/CVF Conference on Computer Vision and Pattern Recognition (CVPR)}, pages 16210--16220, 2022.

\bibitem[Yu et~al.(2021)Yu, Li, Tancik, Li, Ng, and Kanazawa]{plenoctrees}
Alex Yu, Ruilong Li, Matthew Tancik, Hao Li, Ren Ng, and Angjoo Kanazawa.
\newblock Plenoctrees for real-time rendering of neural radiance fields.
\newblock In \emph{2021 {IEEE/CVF} International Conference on Computer Vision, {ICCV} 2021, Montreal, QC, Canada, October 10-17, 2021}, pages 5732--5741. {IEEE}, 2021.

\bibitem[Zavrtanik et~al.(2021)Zavrtanik, Kristan, and Skocaj]{AD:draem}
Vitjan Zavrtanik, Matej Kristan, and Danijel Skocaj.
\newblock Dr{\ae}m - {A} discriminatively trained reconstruction embedding for surface anomaly detection.
\newblock In \emph{2021 {IEEE/CVF} International Conference on Computer Vision, {ICCV} 2021, Montreal, QC, Canada, October 10-17, 2021}, pages 8310--8319. {IEEE}, 2021.

\bibitem[Zhou et~al.(2023)Zhou, Li, Jiang, Wang, Zhou, Zhang, and Zhao]{madsim}
Qiang Zhou, Weize Li, Lihan Jiang, Guoliang Wang, Guyue Zhou, Shanghang Zhang, and Hao Zhao.
\newblock {PAD:} {A} dataset and benchmark for pose-agnostic anomaly detection.
\newblock In \emph{NeurIPS}, 2023.

\end{thebibliography}
}


\clearpage
\setcounter{page}{1}
\setcounter{table}{0}
\maketitlesupplementary

\setcounter{section}{0}
\renewcommand{\thesection}{\Alph{section}}%
\section{Experimental Results}
\label{sec:supp_detection}

We report the full results of our quantitative experiments for all categories in both MAD~\cite{madsim} and the \gls{nerf} synthetic scenes~\cite{NERF}.

\subsection{Anomaly Detection}\label{sec:supplement_AD}

The full remaining results of our anomaly segmentation experiments in~\cref{sec:results} are reported in this section. The pixel-wise AUROC is given in~\cref{tab:supp_pixel-wise} and the AUPRO in~\cref{tab:supp_aupro}. Since achieving higher scores in AUPRO is more difficult than the pixel-wise AUROC, the margins between \methodreg{} and OmniAD are much larger. Still, \methodreg{} outperforms all other methods.

\begin{table*}[b]
    \resizebox{\linewidth}{!}{
    \begin{tabular}{l|c|cc|ccc|cc|cc|cc}
\toprule
  &  \multicolumn{1}{c|}{{\textbf{Feat.Emb.}}} &  \multicolumn{2}{c|}{{\textbf{Memory banks}}}          & \multicolumn{3}{c|}{{\textbf{Student-Teacher}}}  & \multicolumn{2}{c|}{{\textbf{Normal. Flow}}} & \multicolumn{2}{c|}{{\textbf{Synthetic}}} & \multicolumn{2}{c}{{\textbf{View Synthesis}}}  \\   
   & PaDiM     &  CFA & PatchCore & RD4AD & AST & STFPM &  CFlow      & CS-Flow         & SimpleNet  & DR\AE M  & OmniAD   &   \methodreg{}  \\
  \multirow{-3}{*}{Category} & \cite{AD:padim} & \cite{AD:CFA} & \cite{AD:patchcore} & \cite{AD:RD4AD} &  \cite{AD:AST} & \cite{AD:STFPM} & \cite{AD:cflow-ad} & \cite{AD:csflow} & \cite{AD:simplenet} &  \cite{AD:draem} & \cite{madsim} & (ours) \\\midrule
    Gorilla    & 93.0 & 91.4 & 88.4 & 94.8 & 58.1 & 93.8 & 94.7 & 69.2 & 92.0 & 77.7 & \textbf{\underline{99.5}} & \textbf{\underline{99.5}} $\pm$ 0.01  \\
    Unicorn    & 88.8 & 85.2 & 58.9 & 88.8 & 81.0 & 89.3 & 89.9 & 73.1 & 87.9 & 26.0 & \underline{98.2} & \textbf{99.6} $\pm$ 0.02  \\
    Mallard    & 85.9 & 83.7 & 66.1 & 85.6 & 59.1 & 86.0 & 87.3 & 63.3 & 86.2 & 47.8 & \underline{97.4} & \textbf{99.7} $\pm$ 0.00     \\
    Turtle     & 91.8 & 88.7 & 77.5 & 93.7 & 56.3 & 91.0 & 90.2 & 73.8 & 91.5 & 45.3 & \underline{99.1} & \textbf{99.5} $\pm$ 0.01  \\
    Whale      & 90.1 & 87.9 & 60.9 & 90.9 & 54.2 & 88.6 & 89.2 & 64.4 & 90.7 & 55.9 & \underline{98.3} & \textbf{99.5} $\pm$ 0.05  \\
    Bird       & 93.4 & 92.2 & 88.6 & 92.3 & 38.6 & 90.6 & 91.8 & 80.1 & 91.9 & 60.3 & \underline{95.7} & \textbf{99.4} $\pm$ 0.01  \\
    Owl        & 96.3 & 93.9 & 86.3 & 96.3 & 83.7 & 91.8 & 94.6 & 75.2 & 94.3 & 78.9 & \textbf{99.4} & \underline{99.2} $\pm$ 0.03  \\
    Sabertooth & 94.5 & 88.0 & 69.4 & 92.4 & 71.5 & 89.3 & 93.3 & 70.5 & 90.1 & 26.2 & \underline{98.5} & \textbf{99.4} $\pm$ 0.02  \\
    Swan       & 93.2 & 95.0 & 73.5 & 93.5 & 87.9 & 90.8 & 93.1 & 71.8 & 93.2 & 75.9 & \underline{98.8} & \textbf{99.3} $\pm$ 0.02  \\
    Sheep      & 94.8 & 94.1 & 79.9 & 94.5 & 50.5 & 93.2 & 94.3 & 76.2 & 93.4 & 70.5 & \underline{97.7} & \textbf{99.4} $\pm$ 0.01  \\
    Pig        & 95.2 & 95.6 & 83.5 & 96.8 & 72.9 & 94.2 & 97.1 & 79.1 & 96.8 & 65.6 & \underline{97.7} & \textbf{99.8} $\pm$ 0.00     \\
    Zalika     & 87.8 & 87.7 & 64.9 & 89.6 & 55.7 & 86.2 & 89.4 & 65.7 & 89.6 & 66.6 & \underline{99.1} & \textbf{99.3} $\pm$ 0.04  \\
    Phoenix    & 88.3 & 87.0 & 62.4 & 87.6 & 83.6 & 86.1 & 87.3 & 77.7 & 88.9 & 38.7 & \underline{99.4} & \textbf{99.5} $\pm$ 0.00     \\
    Elephant   & 74.1 & 77.8 & 56.2 & 75.2 & 84.1 & 76.8 & 72.4 & 76.8 & 70.7 & 55.9 & \underline{99.0} & \textbf{99.7} $\pm$ 0.00     \\
    Parrot     & 87.7 & 83.7 & 70.7 & 87.2 & 73.8 & 84.0 & 86.8 & 67.0 & 78.7 & 34.4 & \textbf{\underline{99.5}} & \textbf{\underline{99.5}} $\pm$ 0.01  \\
    Cat        & 94.0 & 95.0 & 85.6 & 94.8 & 55.3 & 93.7 & 94.7 & 61.9 & 93.9 & 79.4 & \underline{97.7} & \textbf{99.3} $\pm$ 0.07  \\
    Scorpion   & 90.7 & 92.2 & 79.9 & 93.6 & 82.6 & 90.7 & 91.9 & 72.2 & 89.1 & 79.7 & \underline{95.9} & \textbf{99.3} $\pm$ 0.01  \\
    Obesobeso  & 95.6 & 96.2 & 91.9 & 95.8 & 60.0 & 94.2 & 95.8 & 80.1 & 96.9 & 89.2 & \underline{98.0} &\textbf{99.5} $\pm$ 0.02  \\
    Bear       & 92.2 & 90.7 & 79.5 & 92.8 & 81.0 & 90.6 & 92.2 & 74.9 & 92.3 & 39.2 & \underline{99.3} & \textbf{99.6}$\pm$ 0.00     \\
    Puppy      & 87.5 & 82.3 & 73.3 & 89.5 & 71.6 & 84.9 & 89.6 & 62.0 & 85.5 & 45.8 & \underline{98.8} & \textbf{99.1} $\pm$ 0.03  \\\midrule
    mean         & 90.7	& 89.4 & 74.9 & 91.3 & 	68.1	& 89.3	& 90.8	& 71.7	& 89.7	& 58.0	& \underline{98.4}	& \textbf{99.5}	$\pm$ 0.01
 \\
    \end{tabular}
    }
    \caption{AUROC $(\uparrow)$ for pixel-level anomaly detection performance on MAD. We run our approach n = 5 times and mark the best result in bold and the runner-up underlined}
    \label{tab:supp_pixel-wise}
\end{table*}

\begin{table*}[h]
    \resizebox{\linewidth}{!}{
    \begin{tabular}{l|c|cc|ccc|cc|cc|cc}
\toprule
  &  \multicolumn{1}{c|}{{\textbf{Feat.Emb.}}} &  \multicolumn{2}{c|}{{\textbf{Memory banks}}}          & \multicolumn{3}{c|}{{\textbf{Student-Teacher}}}  & \multicolumn{2}{c|}{{\textbf{Normal. Flow}}} & \multicolumn{2}{c|}{{\textbf{Synthetic}}} & \multicolumn{2}{c}{{\textbf{View Synthesis}}}  \\   
   & PaDiM     &  CFA & PatchCore & RD4AD & AST & STFPM &  CFlow      & CS-Flow         & SimpleNet  & DR\AE M  & OmniAD   &   \methodreg{}  \\
  \multirow{-3}{*}{Category} & \cite{AD:padim} & \cite{AD:CFA} & \cite{AD:patchcore} & \cite{AD:RD4AD} &  \cite{AD:AST} & \cite{AD:STFPM} & \cite{AD:cflow-ad} & \cite{AD:csflow} & \cite{AD:simplenet} &  \cite{AD:draem} & \cite{madsim} & (ours) \\\midrule
    Gorilla    & 76.7 & - & 80.8 & 79.7 & 30.7 & - & - & 27.1 & 75.1  & - & \underline{94.5}          & \textbf{94.6} $\pm$ 0.19  \\
    Unicorn    & 66.8 & - & 81.2 & 74.8 & 37.3 & - & - & 19.8 & 73.1 & - & \underline{85.4}          & \textbf{95.4} $\pm$ 0.06  \\
    Mallard    & 60.5 & - & 75.1 & 62.7 & 33.5 & - & - & 19.6 & 61.0 & - & \underline{81.5 }         & \textbf{97.2} $\pm$ 0.02  \\
    Turtle     & 77.8  & - & \underline{83.5} & 81.6 & 24.2 & - & - & 32.7 & 73.3 & - & 79.7          & \textbf{97.5} $\pm$ 0.2   \\
    Whale      & 76.4 & - & 82.1 & 79.8 & 18.0  & - & - & 20.9 & 82.6 & - & \underline{93.3}          & \textbf{97.5} $\pm$ 0.25  \\
    Bird       & 80.3  & - & \underline{81.4} & 79.7 & 01.0  & - & - & 27.2 & 78.4  & - & 75.4          & \textbf{95.8} $\pm$ 0.08  \\
    Owl        & 88.5 & - & 86.2 & 88.8  & 54.8 & - & - & 30.1  & 83.1 & - & \textbf{95.2} & \underline{94.2}          $\pm$ 0.28  \\
    Sabertooth & \underline{83.8} & - & 80.4  & 75.8 & 45.9 & - & - & 22.5 & 74.9 & - & 83.2          & \textbf{95.4} $\pm$ 0.15  \\
    Swan       & 81.9 & - & 87.7 & 81.7  & 63.1 & - & - & 28.3 & 83.2 & - & \underline{93.0}          & \textbf{96.3} $\pm$ 0.07  \\
    Sheep      & 87.9 & - & \underline{89.0} & 87.6 & 12.8 & - & - & 26.2 & 84.9 & - & 71.9          & \textbf{96.3} $\pm$ 0.07  \\
    Pig        & 81.4 & - & \underline{88.7} & 84.9 & 12.3  & - & - & 27.2 & 86.0 & - & 84.0          & \textbf{96.9}  $\pm$ 0.05  \\
    Zalika     & 71.3 & - & 80.7 & 74.4 & 38.2 & - & - & 31.1 & 76.6 & - & \underline{90.3}         & \textbf{91.3} $\pm$ 0.11  \\
    Phoenix    & 72.0 & - & 81.5 & 74.4 & 61.6 & - & - & 27.9 & 75.9 & - & \underline{93.7}          & \textbf{93.9} $\pm$ 0.12  \\
    Elephant   & 64.6 & - & 74.0    & 67.7 & 67.0 & - & - & 34.3 & 69.8  & - & \underline{91.8}          & \textbf{95.7} $\pm$ 0.1   \\
    Parrot     & 72.1 & - & 84.0 & 72.2 & 45.3 & - & - & 37.2 & 62.6 & - & \underline{94.9}          & \textbf{95.9} $\pm$ 0.04  \\
    Cat        & 85.0 & - & \underline{88.7} & 88.1 & 12.6 & - & - & 22.9 & 85.8 & - & 71.4          & \textbf{94.7} $\pm$ 0.33  \\
    Scorpion   & 77.2 & - & \underline{88.7} & 84.1 & 47.8 & - & - & 21.5 & 76.9 & - & 79.8          & \textbf{96.7} $\pm$ 0.09  \\
    Obesobeso  & 88.2 & - & 90.2 & 90.4 & 41.7 & - & - & 29.9 & \underline{91.3} & - & 79.9          & \textbf{94.8} $\pm$ 0.11  \\
    Bear       & 79.5 & - & 88.8 & 82.1 & 62.4 & - & - & 28.6 & 84.1 & - & \underline{97.0}          & \textbf{97.6} $\pm$ 0.03  \\
    Puppy      & 68.3 & - & 81.0    & 73.6 & 42.4 & - & - & 27.0 & 68.8 & - & \underline{96.1}          & \textbf{97.4} $\pm$ 0.08  \\\midrule
    mean         & 77.0    & - & 83.7 & 79.2 & 37.6 & - & - & 27.1 & 77.4 & - & \underline{86.6 }          & \textbf{95.8} $\pm$ 0.03 
    \end{tabular}
    }
    \caption{AUPRO $(\uparrow)$ for anomaly segmentation performance on MAD. We run our approach n = 5 times and mark the best result in bold and the runner-up underlined. We only report AUPRO for the methods reproduced by us.}
    \label{tab:supp_aupro}
\end{table*}

\FloatBarrier
\subsection{Pose Estimation}\label{sec:supplement_pose_estimation}

We report the full results of our pose estimation experiments from~\cref{sec:4_poseest} on the \gls{nerf} synthetic scenes~\cite{NERF} in~\cref{tab:supp_pose_synth} and on \gls{mad}~\cite{madsim} in~\cref{tab:supp_pose_MAD}. For both translation and rotation, we \methodreg{} beats iNeRF by clear margins. It should also be noted, that we match the rotation up to a thousandth of a radian in most of the categories, showing the precision of our pose estimation.

\begin{table*}[b]
    \centering
    \begin{tabular}{l|ccc||ccc}
    \toprule
       Category &  \multicolumn{3}{c||}{Translation Error (total) $\downarrow$ } & \multicolumn{3}{c}{Rotation Error (rad) $\downarrow$ }\\\cline{2-7}
       (\gls{nerf}) & Coarse~\cite{madsim} & iNeRF~\cite{iNERF} & \methodreg{} & Coarse~\cite{madsim} & iNeRF~\cite{iNERF} & \methodreg{}\\\midrule
        chair & 0.548 & 0.033 & \textbf{0.002} & 0.074 & 0.004 & \textbf{0.000}\\
        drums & 0.616 & 0.018 & \textbf{0.007} & 0.083 & 0.002 & \textbf{0.000}\\
        ficus & 0.805 & 0.085 & \textbf{0.055} & 0.111 & 0.011 & \textbf{0.008} \\
        hotdog & 0.561 & 0.041 & \textbf{0.013}& 0.075 & 0.005 & \textbf{0.001}  \\
        lego & 0.561 & 0.014 & \textbf{0.003}  & 0.076 & 0.002 & \textbf{0.000}\\
        materials & 0.709 & 0.071 & \textbf{0.035} & 0.097 & 0.010 & \textbf{0.004}  \\
        mic & 0.588 & 0.135 & \textbf{0.042} & 0.079 & 0.017 & \textbf{0.005} \\
        ship & 0.606 & 0.045 & \textbf{0.012} & 0.082 & 0.005 & \textbf{0.001}\\\midrule
        mean & 0.624 & 0.055 & \textbf{0.021} & 0.084 & 0.007 & \textbf{0.003}\\
    \end{tabular}
    \caption{Error in Translation and Rotation on the \gls{nerf}  synthetic scenes~\cite{NERF}. Best results in \textbf{bold}.}
    \label{tab:supp_pose_synth} 
\end{table*}

\begin{table*}[b]
\centering
    \begin{tabular}{l|ccc||ccc}
    \toprule
       Category &  \multicolumn{3}{c||}{Translation Error (total) $\downarrow$ }  & \multicolumn{3}{c}{Rotation Error (rad) $\downarrow$ } \\\cline{2-7}
       (MAD) & Coarse~\cite{madsim} & iNeRF~\cite{iNERF} & \methodreg{} & Coarse~\cite{madsim} & iNeRF~\cite{iNERF} & \methodreg{} \\\midrule
        Gorilla    & 0.790 & 0.029 & \textbf{0.006} & 0.131 & 0.010 & \textbf{0.000} \\
        Unicorn    & 0.880 & 0.156 & \textbf{0.006} & 0.122 & 0.025 & \textbf{0.000}\\
        Mallard    & 0.642 & 0.101 & \textbf{0.003} & 0.120 & 0.018 & \textbf{0.000} \\
        Turtle     & 0.649 & 0.112 & \textbf{0.004} & 0.127 & 0.025 & \textbf{0.000}  \\
        Whale      & 0.797 & 0.091 & \textbf{0.005} & 0.130 & 0.014 & \textbf{0.001} \\
        Bird       & 0.786 & 0.272 & \textbf{0.185}   & 0.197 & 0.089 & \textbf{0.073} \\
        Owl        & 1.433 & 0.914 & \textbf{0.871}  & 0.492 & 0.413 & \textbf{0.412} \\
        Sabertooth & 0.671 & 0.072 & \textbf{0.004}  & 0.128 & 0.021 & \textbf{0.001} \\
        Swan       & 0.774 & 0.189 & \textbf{0.007}  & 0.133 & 0.041 & \textbf{0.001}\\
        Sheep      & 1.105 & 0.528 & \textbf{0.449}  & 0.342 & \textbf{0.206} & 0.224 \\
        Pig        & 0.687 & 0.043 & \textbf{0.004} & 0.126 & 0.007 & \textbf{0.000}  \\
        Zalika     & 0.690 & 0.124 & \textbf{0.004} & 0.126 & 0.025 & \textbf{0.000}\\
        Phoenix    & 0.851 &  \textbf{0.146} & 0.148 & 0.154 & \textbf{0.037} & \textbf{0.037}  \\
        Elephant   & 0.818 & 0.285 & \textbf{0.142} & 0.155 & 0.079 & \textbf{0.037} \\
        Parrot     & 0.797 & 0.178 & \textbf{0.005} & 0.131 & 0.036 & \textbf{0.000}  \\
        Cat        & 0.684 & 0.051 & \textbf{0.007} & 0.134 & 0.009 & \textbf{0.001} \\
        Scorpion   & 0.743 & 0.086 & \textbf{0.006} & 0.136 & 0.015 & \textbf{0.001} \\
        Obesobeso  & 0.677 & 0.016 & \textbf{0.006} & 0.118 & 0.003 & \textbf{0.001} \\
        Bear       & 0.723 & 0.054 & \textbf{0.006} & 0.126 & 0.017 & \textbf{0.001} \\
        Puppy      & 0.863 & 0.130 & \textbf{0.006} & 0.135 & 0.031 & \textbf{0.001}  \\\midrule
        mean       & 0.803 & 0.179 & \textbf{0.094}  & 0.163 & 0.056 & \textbf{0.040}\\ 
    \end{tabular}
    \caption{Error in Translation and Rotation on the MAD data set~\cite{madsim}. Best results in \textbf{bold}.}
    \label{tab:supp_pose_MAD}    
\end{table*}

\clearpage
\subsection{Sparse-View Data}\label{sec:supplement_sparse}

We present the full results of our sparse-view data experiments on \gls{mad}~\cite{madsim} from~\cref{sec:4_sparse}. We report the image-wise results in ~\cref{tab:supp_sparse_auroc}. For the segmentation task, the pixel-wise AUROCS are given in ~\cref{tab:supp_sparse_pixels}, and the AUPROS in ~\cref{tab:supp_sparse_aupros}.

With \glspl{nerf} struggling in sparse-view settings, \methodreg{} decisively beats OmniAD for all steps of view-sparsification. Fewer views also result in larger margins.

\begin{table*}[b]
\centering
    \begin{tabular}{l|cc|cc|cc|cc}
    \toprule
       \multicolumn{1}{c|}{Sparsity} &  \multicolumn{2}{c|}{$20\%$} &  \multicolumn{2}{c|}{$40\%$} &  \multicolumn{2}{c|}{$60\%$} &  \multicolumn{2}{c}{$80\%$} \\\hline
       \multicolumn{1}{c|}{Category} & OmniAD & \methodreg{} & OmniAD & \methodreg{} & OmniAD & \methodreg{}  & OmniAD & \methodreg{} \\\midrule
        Gorilla    & 71.0 & \textbf{79.1} & 75.9          & \textbf{83.7} & 82.7 & \textbf{85.8} & \textbf{93.0} & 86.6           \\
        Unicorn    & 81.6 & \textbf{91.6} & 87.1          & \textbf{98.2} & 87.6 & \textbf{98.8} & 85.0          & \textbf{98.5}  \\
        Mallard    & 80.8 & \textbf{90.6} & 82.1          & \textbf{95.4} & 86.1 & \textbf{96.7} & 86.6          & \textbf{96.7}  \\
        Turtle     & 62.5 & \textbf{66.5} & 98.4          & \textbf{97.2} & 88.9 & \textbf{97.0} & 79.5          & \textbf{97.4}  \\
        Whale      & 57.1 & \textbf{78.8} & 76.7          & \textbf{93.5} & 81.1 & \textbf{92.4} & \textbf{92.8} & 91.4           \\
        Bird       & 65.0 & \textbf{73.8} & 77.4          & \textbf{88.5} & 81.3 & \textbf{91.6} & 74.9          & \textbf{94.8}  \\
        Owl        & 67.2 & \textbf{68.7} & \textbf{73.2} & 72.3          & 83.9 & \textbf{87.8} & \textbf{94.9} & 84.3           \\
        Sabertooth & 74.2 & \textbf{76.4} & 93.5          & \textbf{92.7} & 95.0 & \textbf{94.4} & 91.9          & \textbf{96.8}  \\
        Swan       & 67.0 & \textbf{84.7} & 80.8          & \textbf{90.2} & 78.3 & \textbf{89.1} & 92.6          & \textbf{93.0}  \\
        Sheep      & 61.5 & \textbf{75.2} & 74.0          & \textbf{92.0} & 70.5 & \textbf{88.6} & 73.1          & \textbf{93.6}  \\
        Pig        & 68.6 & \textbf{77.1} & 65.4          & \textbf{88.3} & 78.5 & \textbf{95.9} & 84.1          & \textbf{96.3}  \\
        Zalika     & 68.2 & \textbf{75.8} & 79.7          & \textbf{82.1} & 81.7 & \textbf{86.9} & 88.3          & \textbf{89.2}  \\
        Phoenix    & 66.6 & \textbf{68.0} & 77.1          & \textbf{78.3} & 76.7 & \textbf{83.6} & \textbf{92.6} & 81.7           \\
        Elephant   & 68.0 & \textbf{68.1} & 90.0          & \textbf{88.4} & 93.3 & \textbf{93.4} & 88.9          & \textbf{96.8}  \\
        Parrot     & 67.7 & \textbf{76.7} & 84.3          & \textbf{88.2} & 84.1 & \textbf{93.2} & \textbf{96.9} & 96.3  \\
        Cat        & 51.6 & \textbf{72.0} & 51.6          & \textbf{81.0} & 61.6 & \textbf{83.3} & 72.4          & \textbf{83.5}  \\
        Scorpion   & 63.8 & \textbf{83.4} & 76.3          & \textbf{92.9} & 83.7 & \textbf{98.3} & 79.0          & \textbf{99.4}  \\
        Obesobeso  & 56.2 & \textbf{79.3} & 61.7          & \textbf{91.0} & 63.2 & \textbf{92.8} & 79.1          & \textbf{92.8}  \\
        Bear       & 78.1 & \textbf{85.0} & 93.9          & \textbf{98.4} & 98.5 & \textbf{98.7} & 96.6          & \textbf{98.1}  \\
        Puppy      & 66.2 & \textbf{81.2} & 82.6          & \textbf{93.9} & 87.7 & \textbf{95.6} & 92.8          & \textbf{94.2}  \\\midrule
        mean         & 67.2 & \textbf{77.6} & 79.1          & \textbf{89.3} & 82.2 & \textbf{92.2} & 86.8          & \textbf{93.1} 
    \end{tabular}
    \caption{All AUROC scores $(\uparrow)$ measuring the image-wise anomaly detection performance for the sparse-view on \gls{mad}. Best results in \textbf{bold}.}
    \label{tab:supp_sparse_auroc}    
\end{table*}

\begin{table*}[b]
\centering
    \begin{tabular}{l|cc|cc|cc|cc}
    \toprule
       \multicolumn{1}{c|}{Sparsity} &  \multicolumn{2}{c|}{$20\%$} &  \multicolumn{2}{c|}{$40\%$} &  \multicolumn{2}{c|}{$60\%$} &  \multicolumn{2}{c}{$80\%$} \\\hline
       \multicolumn{1}{c|}{Category} & OmniAD & \methodreg{} & OmniAD & \methodreg{} & OmniAD & \methodreg{}  & OmniAD & \methodreg{} \\\midrule
        Gorilla    & 95.8 & \textbf{97.6} & 98.4 & \textbf{99.1} & 98.9 & \textbf{99.4} & 99.3          & \textbf{99.4}  \\
        Unicorn    & 93.6 & \textbf{97.8} & 96.7 & \textbf{99.4} & 96.3 & \textbf{99.5} & 97.0          & \textbf{99.5}  \\
        Mallard    & 94.5 & \textbf{98.6} & 95.5 & \textbf{99.6} & 97.6 & \textbf{99.7} & 96.8          & \textbf{99.7}  \\
        Turtle     & 87.9 & \textbf{96.4} & 93.2 & \textbf{99.0} & 92.9 & \textbf{99.5} & 95.0          & \textbf{99.5}  \\
        Whale      & 90.3 & \textbf{96.6} & 96.0 & \textbf{99.3} & 98.4 & \textbf{99.1} & 98.2          & \textbf{99.4}  \\
        Bird       & 92.4 & \textbf{97.3} & 94.2 & \textbf{99.0} & 93.8 & \textbf{99.1} & 95.2          & \textbf{99.4}  \\
        Owl        & 97.0 & \textbf{97.6} & 97.5 & \textbf{97.9} & 99.0 & \textbf{99.1} & \textbf{99.3} & 99.1           \\
        Sabertooth & 92.4 & \textbf{96.9} & 94.8 & \textbf{99.2} & 97.7 & \textbf{99.3} & 97.9          & \textbf{99.4}  \\
        Swan       & 95.7 & \textbf{98.1} & 97.9 & \textbf{98.9} & 98.0 & \textbf{99.2} & 98.5          & \textbf{99.3}  \\
        Sheep      & 92.8 & \textbf{97.7} & 94.1 & \textbf{99.0} & 93.6 & \textbf{98.8} & 93.6          & \textbf{99.3}  \\
        Pig        & 95.2 & \textbf{98.1} & 94.6 & \textbf{99.1} & 96.3 & \textbf{99.7} & 96.9          & \textbf{99.8}  \\
        Zalika     & 95.2 & \textbf{96.7} & 97.7 & \textbf{98.4} & 98.3 & \textbf{99.2} & 98.6          & \textbf{99.3}  \\
        Phoenix    & 96.3 & \textbf{96.6} & 98.4 & \textbf{99.0} & 98.9 & \textbf{99.4} & 99.3          & \textbf{99.4}  \\
        Elephant   & 91.3 & \textbf{93.1} & 97.8 & \textbf{99.0} & 98.7 & \textbf{99.6} & 98.3          & \textbf{99.7}  \\
        Parrot     & 93.3 & \textbf{96.3} & 97.8 & \textbf{98.2} & 98.4 & \textbf{99.4} & \textbf{99.5} & \textbf{99.5}  \\
        Cat        & 92.6 & \textbf{98.4} & 94.2 & \textbf{99.2} & 94.8 & \textbf{99.3} & 93.6          & \textbf{99.3}  \\
        Scorpion   & 90.9 & \textbf{96.3} & 91.6 & \textbf{97.4} & 93.5 & \textbf{99.1} & 94.7          & \textbf{99.3}  \\
        Obesobeso  & 93.8 & \textbf{98.0} & 95.1 & \textbf{99.3} & 94.9 & \textbf{99.3} & 95.5          & \textbf{99.5}  \\
        Bear       & 96.2 & \textbf{98.0} & 98.7 & \textbf{99.5} & 99.1 & \textbf{99.5} & 99.2          & \textbf{99.6}  \\
        Puppy      & 91.7 & \textbf{96.9} & 94.4 & \textbf{98.8} & 96.8 & \textbf{99.1} & 97.8          & \textbf{98.7}  \\\midrule
        mean         & 93.4 & \textbf{97.2} & 95.9 & \textbf{98.9} & 96.8 & \textbf{99.3} & 97.2          & \textbf{99.4} 
    \end{tabular}
    \caption{All AUROC scores $(\uparrow)$ measuring the pixel-wise anomaly segmentation performance for the sparse-view on \gls{mad}. Best results in \textbf{bold}.}
    \label{tab:supp_sparse_pixels}    
\end{table*}

\begin{table*}[b]
\centering
    \begin{tabular}{l|cc|cc|cc|cc}
    \toprule
       \multicolumn{1}{c|}{Sparsity} &  \multicolumn{2}{c|}{$20\%$} &  \multicolumn{2}{c|}{$40\%$} &  \multicolumn{2}{c|}{$60\%$} &  \multicolumn{2}{c}{$80\%$} \\\hline
       \multicolumn{1}{c|}{Category} & OmniAD & \methodreg{} & OmniAD & \methodreg{} & OmniAD & \methodreg{}  & OmniAD & \methodreg{} \\\midrule
        Gorilla    & 80.0 & \textbf{86.8} & 88.2          & \textbf{92.3} & 91.7 & \textbf{94.2} & 93.0          & \textbf{94.2}  \\
        Unicorn    & 72.0 & \textbf{85.5} & 83.4          & \textbf{93.7} & 81.8 & \textbf{94.8} & 85.0          & \textbf{94.8}  \\
        Mallard    & 76.9 & \textbf{92.0} & 80.0          & \textbf{95.8} & 87.9 & \textbf{97.0} & 86.6          & \textbf{97.0}  \\
        Turtle     & 59.8 & \textbf{86.9} & 73.4          & \textbf{95.8} & 75.3 & \textbf{97.5} & 79.5          & \textbf{97.5}  \\
        Whale      & 66.3 & \textbf{87.4} & 85.6          & \textbf{96.6} & 93.1 & \textbf{95.6} & 92.8          & \textbf{97.1}  \\
        Bird       & 70.0 & \textbf{87.2} & 71.1          & \textbf{94.1} & 70.9 & \textbf{94.5} & 74.9          & \textbf{95.7}  \\
        Owl        & 85.6 & \textbf{87.4} & 89.4          & \textbf{90.3} & 92.7 & \textbf{93.4} & \textbf{94.9} & 93.8           \\
        Sabertooth & 72.9 & \textbf{86.1} & 80.3          & \textbf{94.6} & 91.6 & \textbf{94.9} & 91.9          & \textbf{95.3}  \\
        Swan       & 82.2 & \textbf{91.4} & 90.1          & \textbf{94.9} & 90.0 & \textbf{95.3} & 92.6          & \textbf{96.4}  \\
        Sheep      & 70.5 & \textbf{89.4} & 71.6          & \textbf{94.7} & 71.1 & \textbf{93.7} & 73.1          & \textbf{95.9}  \\
        Pig        & 77.5 & \textbf{89.6} & 75.3          & \textbf{94.0} & 81.5 & \textbf{96.7} & 84.1          & \textbf{96.9}  \\
        Zalika     & 74.9 & \textbf{80.2} & 84.0          & \textbf{87.0} & 87.6 & \textbf{90.5} & 88.3          & \textbf{90.7}  \\
        Phoenix    & 77.7 & \textbf{79.5} & 87.8          & \textbf{89.9} & 89.8 & \textbf{93.2} & 92.6          & \textbf{93.0}  \\
        Elephant   & 65.1 & \textbf{72.8} & 85.5          & \textbf{92.2} & 90.1 & \textbf{94.9} & 88.9          & \textbf{95.7}  \\
        Parrot     & 74.4 & \textbf{81.3} & \textbf{90.0} & \textbf{90.0} & 92.4 & \textbf{95.3} & \textbf{96.9}         & 95.7           \\
        Cat        & 70.3 & \textbf{90.7} & 73.0          & \textbf{94.4} & 75.7 & \textbf{94.7} & 72.4          & \textbf{94.8}  \\
        Scorpion   & 66.9 & \textbf{84.6} & 69.2          & \textbf{89.5} & 74.6 & \textbf{95.8} & 79.0          & \textbf{96.8}  \\
        Obesobeso  & 74.8 & \textbf{89.1} & 77.0          & \textbf{93.4} & 76.7 & \textbf{93.8} & 79.1          & \textbf{94.6}  \\
        Bear       & 85.6 & \textbf{91.9} & 94.5          & \textbf{96.9} & 96.2 & \textbf{97.1} & 96.6          & \textbf{97.6}  \\
        Puppy      & 71.4 & \textbf{90.5} & 80.3          & \textbf{96.6} & 88.7 & \textbf{97.2} & 92.8          & \textbf{96.1}  \\\midrule
        mean         & 73.7 & \textbf{86.5} & 81.5          & \textbf{93.3} & 85.0 & \textbf{95.0} & 86.8          & \textbf{95.5} 
    \end{tabular}
    \caption{All AUPRO scores $(\uparrow)$ measuring the anomaly segmentation performance for the sparse-view on \gls{mad}. Best results in \textbf{bold}.}
    \label{tab:supp_sparse_aupros}    
\end{table*}

\end{document}